\documentclass[preprint,12pt]{elsarticle}
\usepackage{amssymb}
\usepackage{amsmath,amsfonts,amssymb}
\usepackage{algorithmic}
\usepackage{algorithm}
\usepackage{array}
\usepackage{booktabs}
\usepackage{multirow}
\usepackage{xcolor}
\usepackage{graphicx}
\usepackage{subfigure}
\usepackage{bbding}
\usepackage{colortbl}
\usepackage{ulem}
\newtheorem{theorem}{Theorem}[section]
\newtheorem{lemma}{Lemma}[section]
\newtheorem{definition}{Definition}[section]
\newtheorem{remark}{Remark}[section]

\journal{Transportation Research Part C}

\begin{document}

\begin{frontmatter}

\title{MNT-TNN: Spatiotemporal Traffic Data Imputation via Compact Multimode Nonlinear Transform-based Tensor Nuclear Norm}

\author[label1,label2]{Yihang Lu} %% Author name
\author[label1]{Mahwish Yousaf \corref{cor}}
\author[label1]{Xianwei Meng \corref{cor}}
\author[label2]{Enhong Chen}

\cortext[cor]{Corresponding author.}
\affiliation[label1]{organization={Institute of Intelligent Machines, Hefei Institutes of Physical Science, Chinese Academy of Sciences},%Department and Organization
            city={Hefei},
            country={China}}
        
\affiliation[label2]{organization={University of Science and Technology of China},%Department and Organization
            city={Hefei},
            country={China}}

%% Abstract
\begin{abstract}
Imputation of random or non-random missing data is a long-standing research topic and a crucial application for Intelligent Transportation Systems (ITS). However, with the advent of modern communication technologies such as Global Satellite Navigation Systems (GNSS), traffic data collection has {introduced new challenges in random missing value imputation and increasing demands for spatiotemporal dependency modelings}. To address these issues, {we propose a novel spatiotemporal traffic imputation method based on a Multimode Nonlinear Transformed Tensor Nuclear Norm (MNT-TNN), which can effectively capture} the intrinsic multimode spatiotemporal correlations and low-rankness of the traffic tensor, represented as location $\times$ location $\times$ time. To solve the nonconvex optimization problem, we design a proximal alternating minimization (PAM) algorithm with theoretical convergence guarantees. We {also} suggest an Augmented Transform-based Tensor Nuclear Norm Families (ATTNNs) framework to enhance the imputation results of TTNN techniques, especially at very high miss rates. Extensive experiments on real datasets demonstrate that our proposed MNT-TNN and ATTNNs can outperform the compared state-of-the-art imputation methods, completing the benchmark of random missing traffic value imputation.
\end{abstract}


% \begin{highlights}
% \end{highlights}

%% Keywords
\begin{keyword}
Spatiotemporal data, Traffic imputation, Tensor nuclear norm, Non-convex optimization
\end{keyword}

\end{frontmatter}



\section{Introduction}
\label{intro}
Spatiotemporal traffic data collected from various sensing systems (e.g., loop detectors and floating cars) is the foundation for various applications and decision-making processes in Intelligent Transportation Systems (ITS).  Multi-dimensional data, such as color images, videos, and time sequences, is ubiquitous.  As typical multi-dimensional data, spatio-temporal traffic data plays a crucial role in ITS, which has attracted wide attention in recent years. Missing data imputation is one of the most important research questions in spatiotemporal data analysis since accurate and reliable imputation can help various downstream applications, including traffic forecasting, traffic control/management \cite{zheng2020gman}, vehicle demand analysis \cite{traffdetect}, and urban planning \cite{wang2023human}.  Nevertheless, the limitations of matrix structures in capturing the complexities of traffic data are becoming evident as data collection methods advance, leading to the rise of spatiotemporal traffic tensor methods as a new focus in the field.
\begin{figure}[t]
% \vskip 0.2in
% \centering
\includegraphics[width=\textwidth]{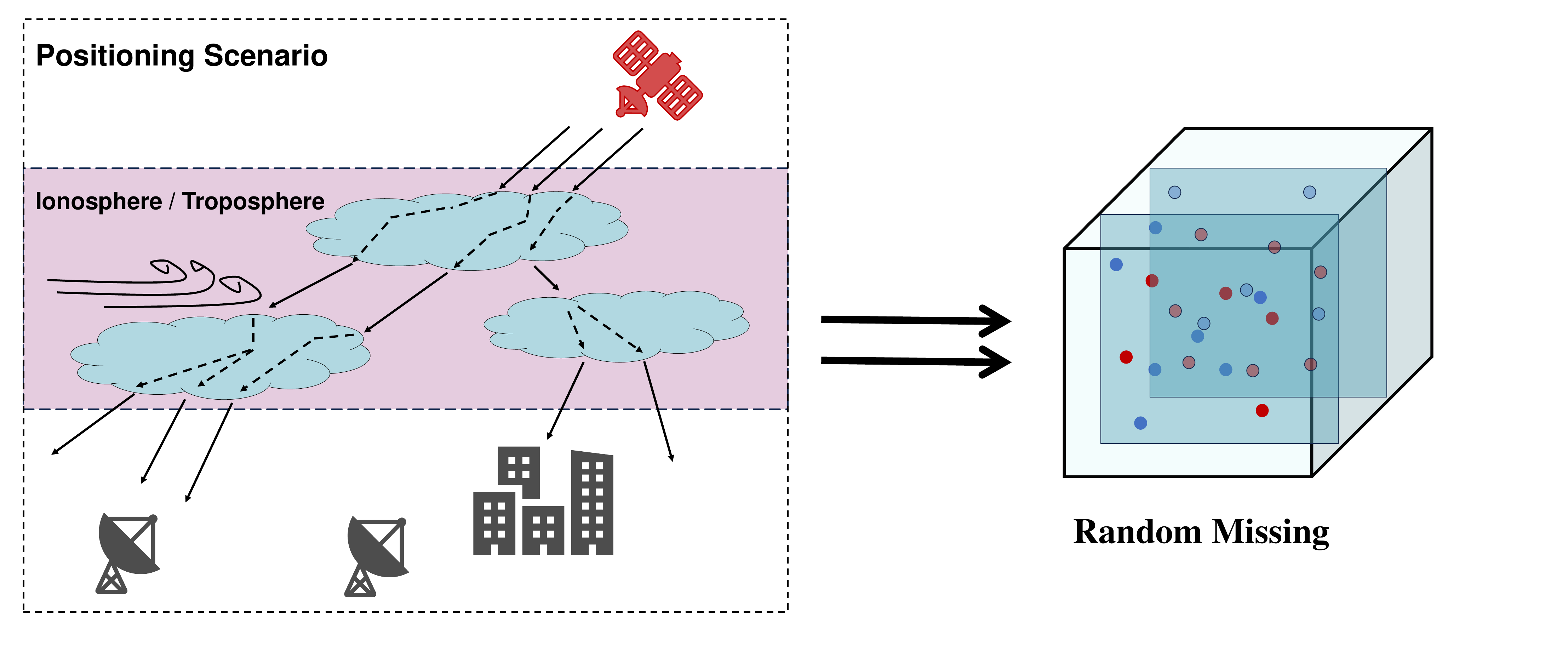}
\caption{Random missing appears to be common in modern data collecting scenarios, raising the need for effective random missing imputation methods for spatiotemporal data.}
\label{fig:motivation1}
% \vskip -0.25in
\end{figure}

\begin{figure*}[htbp]
% \vskip 0.2in
% \centering
\begin{minipage}[t]{\textwidth}
\includegraphics[width=\textwidth]{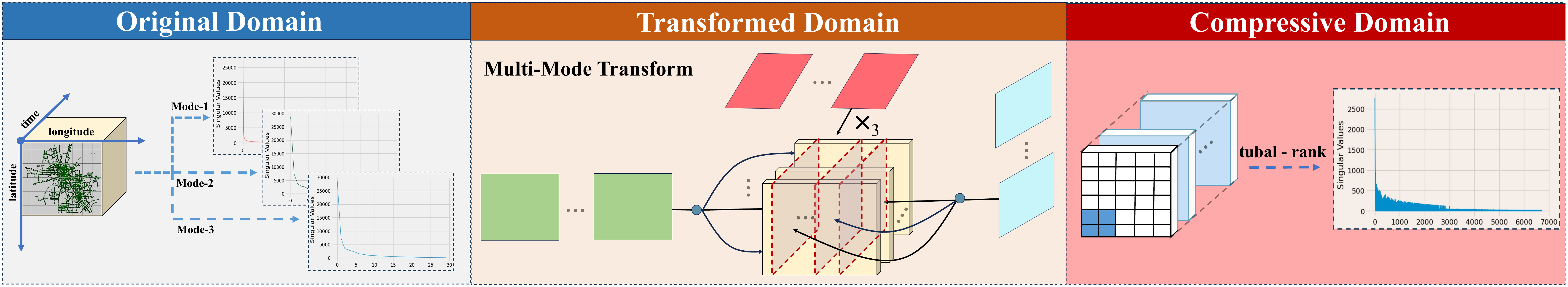}
\end{minipage}
\caption{Multimode nonlinear transform for a unified multimode low-rankness exploration of a spatiotemporal traffic tensor, the sharp decreasing curves of three different modes of the original tensor show their low-rankness respectively, and the transformed core tensor has a low-tubal-rank property.}
\label{fig:motivation2}
% \vskip -0.2in
\end{figure*}

The main challenge with data imputation is effectively understanding and utilizing the intricate relationships and dependencies in both spatial and temporal aspects \cite{wang2018adaptivespatialtemporal}. However, these datasets encounter structural barriers and signal transmission problems during the data collection process, impacting their practicality and efficiency in real-world scenarios. The missing data may not be available because of a malfunctioning sensor, communication error, or maintenance issue. Insufficient sensor coverage in spatial and temporal dimensions is another key factor contributing to the issue of missing data. Incomplete spatiotemporal traffic data consists of multivariate/multidimensional time series with different types and levels of missing values. Essentially, the missing patterns can be summarized into two types: random missing and non-random missing. The former indicates that the missing values occur independently at each position with equal probability, whereas the latter represents missing values that are interrelated and often occur in a connected space. While existing works often emphasize the universality of non-random missing in the field of traffic imputation, it is noteworthy that random missing still occurs in large modern communication techniques like GNSS due to some uncontrollable and chaotic natures ranging from the Atmosphere and Ionosphere, weather change, to outer space circumstances. Therefore, improving data quality and supporting downstream applications requires essential imputation of missing data, and accurately and efficiently imputing large traffic datasets remains a key challenge. Moreover, while Deep Learning (DL) techniques for traffic imputation have become increasingly popular \cite{miao2021generative, tashiro2021csdi, nie2024imputeformer}, they fall short in many practical applications due to the high demands for training data. In other words, without sufficient high-quality training data, deep-learning models often suffer from issues such as instability and poor generalization. In contrast, optimization methods are computationally efficient, and their ability to solve problems online with closed-form solutions makes them robust and adaptable to any volume of data and practical scenarios. Hence, this research focuses on examining optimization-driven random missing imputation techniques, better suited for use on edge devices and in industrial settings.

In the literature, the majority of researchers view traffic data imputation as low-rank approximation issues due to the consistent periodicity and patterns observed in traffic data over time. These characteristics can be effectively described by the concept of low-rankness. Yu et al.\cite{yu2016temporal} suggested including a parameterized autoregressive regularization term in the Low-rank Matrix Factorization (LMF) framework to incorporate temporal dependencies as prior knowledge in traffic imputation. Zhu et al. \cite{zhu2012compressive} used the low-rank hypothesis on a matrix that shows the traffic conditions of road segments at a particular time. They turned the task of filling in missing data into a matrix completion problem by utilizing Probabilistic PCA (PPCA). Qu et al.\cite{qu2009ppca} and Li et al.\cite{li2013efficient} developed an improved model that accurately represents the nonlinear spatiotemporal relationships by utilizing probe vehicle data. Yu et al.\cite{yu2020urban} developed a technique to calculate traffic information for a whole city by employing the Schatten p-norm via matrix completion. Chen et al.\cite{chen2018spatialsvd} introduced the SVD-combined Tensor Decomposition (ScTD) as an enhanced model, integrating multimode biases from traffic patterns and latent features identified by the truncated SVD. They discussed a three-step system suggested for managing incomplete traffic speed information. This technique can identify the traffic trends from data that is only partially observed, and then fill in the missing values. Chen et al.\cite{chen2021latc} proposed a framework called Low-Rank Autoregressive Tensor Completion (LATC) for filling in missing values in spatiotemporal traffic data. Truncated nuclear norm was employed as a useful approximation to circumvent the issue of determining rank in factorization models. Wang et al.\cite{wang2018adaptivespatialtemporal} introduced a novel method for reconstructing traffic data, named Temporal and Adaptive Spatial Constrained Low Rank (TAS-LR). The suggested method utilizes both a low-rank representation model for exploiting a global traffic data feature and an adaptive spatiotemporal constraint for local features.

We tackle the problem of spatiotemporal imputation of traffic data with randomly missing values using the TTNN framework. In previous work, the focus is often on spatiotemporal tensors that are spanned by two temporal dimensions, such as day, times of day \cite{chen2021lstc} or road segment times and time intervals \cite{hu2020robust}. However, different techniques are needed to effectively capture the spatial and temporal variations of a tensor with diverse characteristics. With the advancement of modern communication systems, satellite navigation systems can now collect large quantities of real traffic data, featuring multiple geographical locations with local references that require precise recognition of spatial correlations. It is challenging to use these features effectively to guarantee the accuracy of real-world data with imputation methods. On the other hand, many current studies do not address the imputation of traffic data when faced with very high levels of missing data, an uncommon scenario but still a concern for large-scale traffic datasets\cite{ni2007determiningITS}. 

This paper introduces a new approach called Multimode Nonlinear Transform-based Tensor Nuclear Norm (MNT-TNN) that combines spatial mode data and blends multimode data by integrating a multimode transform within the Transform-based Tensor Nuclear Norm (TTNN) framework. In essence, the initial TTNN requires a linear transform on the third mode of a tensor, and convex/nonconvex optimizations in this framework allow for exploring the Low Tubal Rank (LTR) property, preserving the full structure of the tensor. (Refer to \cite{LuFCLLY20} \cite{ZhangA17} for the LTR definition). Nevertheless, as previously stated, spatiotemporal traffic tensors frequently contain multiple types of features, such as spatial, temporal, and shared spatiotemporal modalities.  As stated by Chen et al. \cite{chen2018spatialsvd}, there may be a hidden implicit low-rankness across multiple modes of the traffic tensor, which must be brought out through specific linear or nonlinear transformations in each mode, as depicted in Fig. \ref{fig:motivation2}. Notably, single-mode TTNN may not be enough to capture such multimode low-rankness, and how to integrate this ability into TTNN remains unknown. To address this problem and maximize the utilization of spatiotemporal relationships in traffic data filling, we first suggest a Multimode Nonlinear Transform (MNT) that combines 1D and 2D transformations related to various modes. Next, we address the nonconvex optimization issue through the application of the Proximal Alternating Minimization (PAM) algorithm with theoretical convergence guarantees. Moreover, we suggest an Augmented TTNN Families (ATTNNs) framework using different TTNN techniques to enhance the imputation performance under very high missing rates. The main contributions of this paper can be summarized as follows:
\begin{itemize}
    \item {We rigorously proposed a novel Multimode Nonlinear Transform-based Tensor Nuclear Norm (MNT-TNN) for the problem of random missing values of spatiotemporal imputation of traffic data.}
    \item We proposed an Augmented Transform-based Tensor Nuclear Norm Families (ATTNNs) framework to improve the imputation results of TTNN techniques, particularly with very high missing rates.
    \item {Extensive experiments are conducted on three real-world datasets to compare the imputation performance of various methods. The results show that MNT-TNN and ATTNNs together achieve superior performance in spatiotemporal traffic imputation across a wide range of missing rates.}
\end{itemize}
The rest of this paper is structured as follows. Section \ref{works} introduces two frameworks that are most related to our method. Section \ref{preliminary} gives some notations and preliminary concepts. Section \ref{method} presents the way for defining our two proposed methods, and provides the solving algorithm as well as proofs for related theorems and properties. Section
\ref{experiment} demonstrates numerical experiments conducted on real data. We provide the potential limitations of this work in Section \ref{limitation} and conclude this work in Section \ref{conclusion}.
\section{Related Works}
\label{works}
\textbf{Low-Rank Tensor Completion for Traffic Imputation:} 
Traditional tensor-based traffic imputation works leverage tensor factorization methods to exploit low-rank properties. For instance, Asif et al. \cite{asif2013cptraffic} used CANDECOMP/PARAFAC (CP) decomposition \cite{yokota2016smoothcp}\cite{erichson2020randomizedcp} to approximate traffic tensors with low CP-rank, while Tan et al. \cite{tan2013tucker}\cite{goulart2017traffictuc} apply Tucker decompositions \cite{gandy2011lownranktucker}, resulting in a core tensor with low Tucker rank. Determining CP-rank of a given tensor is known as an NP-hard problem \cite{hillar2013nphard}, lacking exact algorithms for accurate estimation, while Tucker decomposition needs to factorize the 3D tensor into a core tensor $\mathcal{C}$ and multiple orthogonal factor matrices along each mode,
\begin{gather}
\label{eq:tucker}
    \mathcal{X}=\mathcal{C}\times_1 \textbf{U}_1\times_2 \textbf{U}_2\times \dots \times_d \textbf{U}_d,\\
    \notag{\textbf{U}_i^\top \textbf{U}_i=\textbf{U}_i\textbf{U}_i^\top=I_{m_i\times m_i}},~i=1,2\dots,d
\end{gather}
This process inevitably destroys global correlations within the original tensor, thereby reducing the imputation accuracy. Beyond CP and Tucker decompositions, other tensor decomposition methods, such as Tensor Singular Value Decomposition (TSVD) \cite{nie2023tensorsvd} \cite{deng2021graph}\cite{chen2018spatialsvd} and tensor networks \cite{oseledets2011tensortrain}\cite{zheng2021tensornet}, have been applied to traffic tensor completion with varying degrees of progress. 

As the convex surrogate of the trace of matrix rank, Nuclear Norm (NN) minimization has emerged as an effective and efficient tool to ensure low-rank properties in matrices. Liu et al. \cite{liu2012lrtc} proposed a unified framework referred to as Low-rank Tensor Completion (LRTC), aimed at recovering complete data by minimizing the rank of the observed data tensor: 
\begin{gather}
\label{eq:LRTC}
    \min_{\mathcal{X}} \text{rank}(\mathcal{X})\\
    s.t.~~\mathcal{X}_\Omega = \mathcal{O}_\Omega
\end{gather}
Where $\mathcal{X}\in\mathbb{R}^{n_1\times n_2\times n_3}$ and $\mathcal{O}^{n_1\times n_2\times n_3}$ represent the recovered and observed traffic data, respectively; $\text{rank}(\cdot)$ denotes a well-defined tensor rank function, and $\Omega$ denotes the observed index set. They adapted this framework for visual tensor completion by proposing the HaLRTC model, which solves a low-rank tensor approximation problem using the sum of NNs of unfolded matrices derived from the tensor. This method is easily applicable to spatiotemporal traffic imputation \cite{tan2014trafficlrtc} since no prior knowledge is required. Nonetheless, methods using convex compositions of NNs of unfolded matrices may fall short of fully capturing tensor low-rankness.
To improve this, Chen et al. \cite{chen2020nonconvex} extended non-convex Truncated Nuclear Norm (TNN) minimization to 3D tensors, proposing the Low-Rank Tensor Completion Truncated Nuclear Norm (LRTC-TNN) for enhanced low-rankness exploitation of traffic data. Further, Chen et al. \cite{chen2021lstc} introduced a so-called Low-Tubal-Rank Smoothing Tensor Completion (LSTC-Tubal) method, which incorporates a linear unitary transform into TNN, enabling scalable low-rank tensor processing. Despite these advances, these methods sometimes suffer from over-relaxation due to the nature of NN, which can limit imputation accuracy. To address this problem, Nie et al. \cite{nie2022schattenp} suggested replacing NN with a truncated Schatten p-norm, achieving improved traffic imputation performance.

Although these LRTC methods have yielded strong results in spatiotemporal traffic imputation, existing approaches mainly focus on forming the convex or non-convex problem using various matrix norms and regularization terms. Developing a more compact tensor method specifically tailored for real-world spatiotemporal traffic imputation still deserves exploration.

\textbf{Transform-based TNN Recovery:}
In addition to the LRTC, another paradigm dedicated to low-rank tensor recovery is built upon the transform-based tensor product (t-product) \cite{kernfeld2015tensorprod}, which has been extensively evaluated in the domain of image and video restoration \cite{ZhangA17} because of its conceptual clarity and efficient representation. Literarily, Kilmer et al. \cite{kilmer2021tenalgebra} strictly derive a series of arithmetic operations and theorems based on the t-product, including a tensor-based SVD (t-SVD) and the extension of the Eckart-Young theorem to tensors. The key ingredient for its wide application is the Transform-based Tensor Nuclear Norm (TTNN), which has been proven as the tightest convex envelope of the $l_1$ norm of tensor multi-rank\cite{LuFCLLY20}; and more importantly, TTNN exhibits a very efficient form for supporting both numerical computing and algorithm development. To this end, Zhang et al. \cite{ZhangA17} modeled tensor completion as a convex optimization problem using Fourier transform-based TNN, and in the meanwhile, theoretically extended the Low-rank Matrix Recovery (LRMR) theorem \cite{DBLP:journals/cacm/CandesR12} to tensors. Furthermore, Song et al. \cite{SongNZ20} generalized the method to any unitary transform, and proposed Unitary Transform-based Tensor Nuclear Norm (UTTNN), enabling a robust tensor recovery problem and introducing a new recovery theorem. Subsequently, Jiang et al. \cite{JiangNZH20} proposed a Framelet representation of Tensor Nuclear Norm named FTTNN with its corresponding convex problem. Wang et al. \cite{wang2022conotcoupled} presented a coupled transform-based TNN to extract both spatial and temporal dependencies. Moreover, Li et al. \cite{LiZJZH22} introduced a nonlinear activation function into TTNN, by which a Nonlinear TTNN optimization problem named NTTNN was proposed and significantly improved tensor recovery performance.
\begin{table*}[htbp]
  \setlength{\abovecaptionskip}{0cm}
  \setlength{\belowcaptionskip}{0cm}
  \vspace{-5pt}
\caption{{Comparison between our work and other related methods.}}
  \label{tab:Compare}
  \vspace{5pt}
  \centering
  % \begin{small}
  \renewcommand{\multirowsetup}{\centering}
  \renewcommand\arraystretch{1.1}
  \begin{tabular}{l|cccc}
\toprule
\multirow{2}{*}{Method} & \multicolumn{4}{c}{Feature}  \\ 
\cmidrule(lr){2-5}
& Transformed & Nonlinear & Multimode & Compact \\
\midrule
UTNN \cite{SongNZ20} & \CheckmarkBold & \XSolidBrush & \XSolidBrush & \CheckmarkBold \\
NTTNN \cite{LiZJZH22} & \CheckmarkBold & \CheckmarkBold & \XSolidBrush & \CheckmarkBold \\
LRTC-TNN \cite{chen2020nonconvex} & \XSolidBrush & \XSolidBrush & \CheckmarkBold & \XSolidBrush \\
LSTC-Tubal \cite{chen2021lstc} & \CheckmarkBold & \XSolidBrush & \XSolidBrush & \CheckmarkBold \\
\rowcolor{lightgray}
MNT-TNN (Ours) & \CheckmarkBold & \CheckmarkBold & \CheckmarkBold & \CheckmarkBold \\
\bottomrule
  \end{tabular}
  % \end{small}
  \vspace{-5pt}
\end{table*}

Despite some obscure overlaps with the LRTC methods in traffic imputation, the TTNN methods primarily focused on image and video recovery. Their potencies for imputation of traffic data with random missing values remain largely unexplored. {In Table. \ref{tab:Compare}, we compare our work with these related methods across four dimensions. Here,} {the term 'compact' refers to the model's ability to capture complex multimode correlations within a unified optimization problem, in stark contrast to methods that require separate, sequential steps or treat different data modes independently.}

\section{Preliminaries}
\label{preliminary}
\subsection{Notations}
We will now go over some notations used in this paper. A spatiotemporal graph is given by $\mathcal{G}=(V, E, T)$, where $V, T,$ and $E$ denote the set of vertices, edges, and timestamps of the graph $G$, respectively. Matrices are represented by boldface capital letters like $\mathbf{M}\in{\mathbb{R}^{m\times n}}$ whose $(i,j)$-th element is denoted as $M_{i,j}$. Vectors are denoted by letters in lowercase, such as $v\in \mathbb{R}^m$. Boldface Euler letters, like $\mathcal{X}\in \mathbb{R}^{n_1\times n_2\times n_3}$, are used to represent order-3 tensors. The mode-$k$ matricization (unfolding) of the tensor $\mathcal{X}$ and the inverse operation are denoted as $\mathbf{X}_{(k)},k\in\{n_1,n_2,n_3\}$ and $\text{Fold}(\mathbf{X}_{(k)})$, respectively; and we denote $\mathbf{X}^{(i)}\in \mathbb{R}^{n_1\times n_2}$ to be its $i$-th frontal slice. The Frobeniues norm of $\mathcal{X}$ is defined as $\Vert\mathcal{X}\Vert_F=\sqrt{\sum_{i,j,k}\mathcal{X}_{i,j,k}^2}$ . The nuclear norm and the Frobenius norm of a matrix $\mathbf{X}$ are denoted as $\Vert\mathbf{X}\Vert_*$ and $\Vert \mathbf{X}\Vert_F$, respectively.
\subsection{Key Conceptions of TTNN minimization}
We introduce several key definitions and concepts related to the TTNN framework to provide a solid foundation for our proposed method.

Primarily, TTNN concerns the low-rank properties of 3D tensors $\mathcal{X}^{n_1\times n_2\times n_3}$ within the transformed domain, which indicates that the study focus of this framework is shifted from the original tensor to one transformed along its third mode, i.e., $\mathcal{X}_\mathbf{U}=\mathcal{X} \times_3 \mathbf{U} = \text{Fold}_3(\mathbf{UX}_{(3)})$, where $\mathbf{U}\in \mathbb{R}^{n_3 \times k}$ denotes the factor of transform. Since the traffic imputation in this paper relies on unitary transform-based random recovery theorems, all transform factors used in the following definitions and statements are assumed to be unitary transforms.    
\begin{definition}[Transformed Tensor Multi-rank (TTMR) \cite{ZhangA17}]
\label{def:tmr}
    The TTMR of a tensor $\mathcal{X} \in \mathbb{C}^{n_1\times n_2\times n_3}$ with respect to a transform $\textbf{U}$ is a vector whose $i$-th entry denotes the rank of the $i$-th frontal slice of the tensor $\mathcal{X}_\mathbf{U}$.
\end{definition}
\begin{definition}[Transformed Tensor Nuclear Norm \cite{SongNZ20}]
    The TTNN of a tensor $\mathcal{X} \in \mathbb{C}^{n_1\times n_2\times n_3}$ with respect to a transform $\textbf{U}$  is defined as the sum of the nuclear norms of all frontal slices of the tensor in the transformed domain,
    \begin{equation}
        \Vert \mathcal{X}\Vert_\text{TTNN} = \sum_{i=1}^{n}\Vert{\mathcal{X}_\mathbf{U}}^{(i)}\Vert_*
    \end{equation}
\end{definition}
It can be proved that TTNN is the convex envelope of the $l_1$ norm of the transformed multi-rank (def. \ref{def:tmr}). Accordingly, a Bernoulli random sampling tensor recovery theorem is given as follows,
\begin{theorem}
\label{the:2}
Under certain mild conditions stated in \cite{SongNZ20}, and with the observation set $\Omega$ being uniformly distributed among all sets of cardinality $m=\rho n_1n_2n_3.$ Also, suppose that each observed entry is independently corrupted with probability $\gamma$. Then,there exist universal constants $c_{1},c_{2}>0$ such that with probability at least $1- c_1( n_{( 1) }n_3) ^{- c_2}$, the recovery of $\mathcal{X}$ with $\lambda = 1/ \sqrt {\rho n_{( 1) }n_3}$ is exact, provided that
\begin{equation}
    r\leq\frac{c_rn_{(2)}}{\mu(\log(n_{(1)}n_3))^2}\quad\text{and}\quad\gamma\leq c_\gamma
\end{equation}
where $\rho$ denotes the sampling ratio, $c_r$ and $c_\gamma$ are two positive constants.
\end{theorem}
The core assertion of this theorem is that, under the scheme of unitary transform-based TNN, the completion of a low-rank tensor is exact with an overwhelming probability, provided that the sampling ratio is sufficiently high.
\section{Methodology}
\label{method}
We address the problem of spatiotemporal imputation of traffic data with random missing values based on the TTNN framework. To align with our objectives and to formally establish the proposed optimization problem, we begin by extending some fundamental tensor operations, including mode unfolding and mode product. For simplicity, the following definitions are presented in the context of third-order real tensors. However, these operations can be easily generalized to higher-dimensional complex tensors.
\begin{definition}[Generalized Mode Unfolding (GMU)]
Given a 3D real tensor $\mathcal{X}\in \mathbb{R}^{n_1\times n_2\times n_3}$, the GMU is given as
\begin{equation}
\label{eq:2}
    \mathbf{Unfold}(\mathcal{X},S):=\mathbf{X}_{[S]} \in \mathbb{R}^{{\times}_{i\in S} n_i{\times} (\prod_{j\in N,j\notin S} n_j)}
\end{equation}
where $S$ is an ordered subset of the indices's set $N = \{1,2,3\}$. The symbol ${\times}$ denotes the Cartesian product, which should be distinguished from the product of numbers represented by $\prod$.
The inverse operation of unfolding is represented by  
\begin{equation}
\mathbf{Fold}_{S}(X_{[S]}) := \mathcal{X}
\end{equation}
In addition, we can define a variant of this unfolding,
\begin{equation}
\label{eq:4}
\mathbf{\overline{X}}_{[S]}\in \mathbb{R}^{\prod_{i\in S}n_i{\times} \prod_{j\in N,j\notin S}n_j}
\end{equation}
\end{definition}
{For an exact example, by choosing $S = \{2\}$ and $\{2,3\}$, the tensor $\mathcal{X}$ can be unfolded respectively as a matrix $\mathbf{X}_{[2]}\in\mathbb{R}^{n_2\times n_1n_3}$ and a tensor $\mathbf{X}_{[\{2,3\}]}\in\mathbb{R}^{n_2\times n_3 \times n_1}$ through Eq.(\ref{eq:2}). If use Eq.(\ref{eq:4}), the results will be a matrix $\overline{{\mathbf{X}}}_{[2]}\in\mathbb{R}^{n_2\times n_1n_3}$ and a tensor $\overline{\mathbf{X}}_{[\{2,3\}]}\in\mathbb{R}^{n_2n_3\times n_1}$.}

\begin{definition}[2D Mode Product]
Suppose a real matrix $\mathbf{M}\in \mathbb{R}^{m \times n_k}$, recall that the mode-$k$ product of tensor $\mathcal{X}$ with respect to $\mathbf{M}$ is defined as  $\mathcal{X} \times_k \mathbf{M}=\mathbf{Fold}_{(k)}(\mathbf{M\mathbf{X}}_{(k)}),~k\in N$. In an analogous way, we define a 2D mode product for an arbitrary tensor with respect to any linear operator, involving two similar algebraic operations as follows:
    
By setting $C(S)=2$ in Eq. (\ref{eq:2}), we define the mode-($k,p$) product of tensor $\mathcal{X}$ with respect to the matrix $\mathbf{M}\in \mathbb{R}^{m\times n_k}$ as the form of the so-called face-wise product as 
\begin{gather}
\label{eq:2dtrans}
(\mathcal{X}{\star}_{(k,p)} \mathbf{M})^{(i)}:=\mathbf{M}\mathbf{X}_{[(k,p)]}^{(i)},\\
\notag k,p~\in N,i=1,2,\dots,\prod \limits_{j\in N,j\neq k,p}n_j
\end{gather}
\begin{remark}
    One can notice that when $C(S)=1$, it degenerates to the normal mode-$k$ product after removing the superscript $(i)$ and changing $\star$ into $\times$. Notably, for 3D tensors, the effect of applying this product is equivalent to the mode-$k$ product.
\end{remark} 

Let the factor matrix $\mathbf{M}$ belong to $\mathbb{R}^{m\times n_kn_p}$. The variant of this product is defined by first applying an additional vectorization to each slice to be multiplied,
\begin{equation}
\mathcal{X}~{\bar\star}_{(k,p)}\mathbf{M}:=\mathbf{Fold}_{(k,p)}(\mathbf{M}\overline{\mathbf{X}}_{[(k,p)]})
\end{equation}
\end{definition}
\begin{remark}
These products are based entirely on variations in shape and linear algebra within the vector space, ensuring that they are well-defined. Analogous to the mode-$k$ product, which acts as a 1D transformation applied to each fiber-tube along the $k$-th mode, the 2D mode product in Eq. (\ref{eq:2dtrans}) is precisely a transformation applied to each frontal slice along a specified mode.
\end{remark}
\begin{definition}[Multimode Nonlinear Transform (MNT)]
 For any tensor $\mathcal{X}\in \mathbb{R}^{n_1\times n_2\times n_3}$, the MNT is defined as 
\begin{equation}
\label{eq:transform}
\mathcal{C} = \psi(\mathcal{X}~\bar\star_{p\in P} \mathbf{U}_p \star_{q\in Q} \mathbf{U}_q),~~P,Q\subseteq 2^N
\end{equation}
where $\psi(\cdot)$ is a specified element-wise function, $2^N$ is the power set of $N$, and $P, Q$ are ordered subsets. Unlike the Tucker decomposition described in Eq. (\ref{eq:tucker}), here $\mathbf{U}_p$ and $\mathbf{U}_q$ can be relaxed to semi-orthogonal matrices. We refer to the resulting tensor $\mathcal{C}$ as the transformed kernel/core of the original tensor. 
\end{definition}
\begin{figure*}[t]
% \vskip 0.2in
\centering
\subfigure{
\begin{minipage}[t]{\linewidth}
\includegraphics[width=\linewidth]{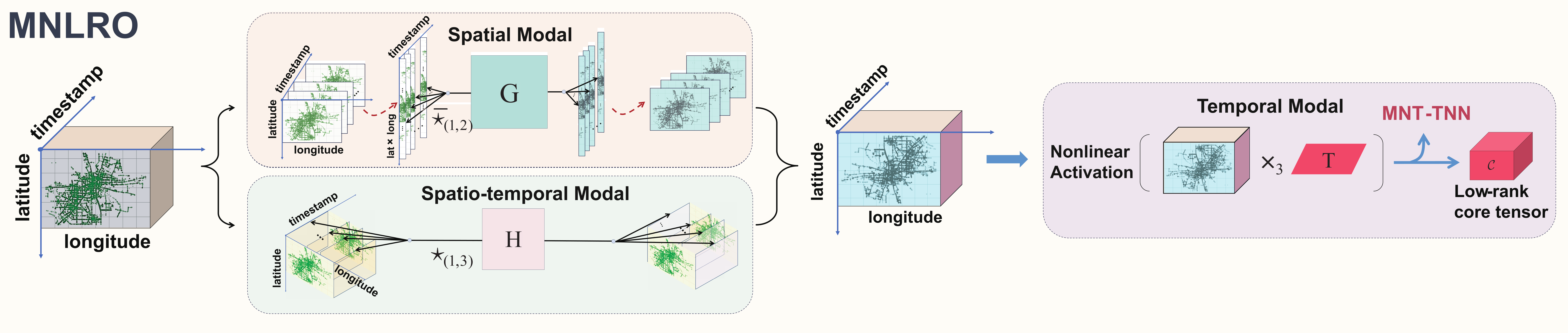}
\end{minipage}
}
\label{Fig.2}
\caption{The flowchart of the Multimode Nonlinear Low-Rank Optimization (MNLRO).}
\vskip -0.2in
\end{figure*}
\begin{definition}[Multimode Nonlinear Transform-based Tensor Nuclear Norm]
For any tensor $\mathcal{X}\in \mathbb{R}^{n_1\times n_2\times n_3}$, its MNT-TNN, denoted as $\Vert\mathcal{X}\Vert_\text{MNT-TNN}$, is defined as follows,
\begin{gather}
\Vert\mathcal{X}\Vert_{\text{MNT-TNN}} = \sum_{i=1}^{m_3}\Vert \mathbf{C}^{(i)}\Vert_*,\\ \nonumber\mathcal{C} = \psi(\mathcal{X}~\bar\star_{p\in \mathbf{P}}\mathbf{U}_p \star_{q\in \mathbf{Q}}\mathbf{U}_q)
\end{gather}
{where $m_3$ denotes the length of the third dimension of the core tensor $\mathcal{C}\in\mathbb{R}^{m_1\times m_2\times m_3}$.} 
\end{definition}
Unlike TNNs, which are restricted to a single-mode linear transform \cite{ZhangEAHK14}, we extend them in this work by applying our multi-mode transform.

{Further observations regarding the preceding definitions can strengthen their conceptual understanding and expand their range of application. The technical details are provided in \ref{app:tq}.} Now, we introduce the following lemma:
\begin{lemma}
\label{lemma:1}
Denoting $\|\mathcal{X}\|_\text{TTNN}= \sum_{i=1}^{n_3} \|\mathbf{X}^{(i)}_\mathbf{U}\|_*$ as the transformed nuclear tensor norm of an order-3 tensor $\mathcal{X} \in \mathbb{C}^{n_1\times n_2\times n_3}$ where $\mathcal{X}_\mathbf{U} \in \mathbb{C}^{n_1\times n_2\times n_3}=\text{Fold}_3(\mathcal{X}\times_3 \mathbf{U})$ and $\mathbf{U}$ is a unitary matrix, then it is the convex envelope of the $l_1$ norm of TMR (see definition \ref{def:tmr}) over a unit ball of the tensor spectral norm.
\end{lemma}
\textit{proof:} The proof of lemma \ref{lemma:1} can be found in Appendix A of the reference paper \cite{SongNZ20}.

Then, the following observation can be made:
\begin{theorem}
\label{theorem}
MT-TNN, namely the MNT-TNN without the nonlinear activation, is the convex envelope of the sum of the TRM (see definition \ref{def:tmr}) of the tensor in a composite transformed domain. 
\end{theorem} In light of the existing proof about TTNN, the proof for this theorem is straightforward, 
By Lemma \ref{lemma:1}, we prove Theorem \ref{theorem} using reduction. 
It suffices to show that for any order-3 real tensor $\mathcal{X}\in\mathbb{R}^{n_1\times n_2\times n_3}$, there exists an surrogate tensor $\mathcal{Z}$ and an orthogonal transformation factor $U$ such that 
\begin{align*}
\|\mathcal{X}\|_\text{MNT-TNN}=\sum_{i=1}^{n_3}\|(\mathcal{X}~\bar\star_{p\in P}\mathbf{U}_p \star_{q\in Q} \mathbf{U}_q )^{(i)}\|_*\\=\sum_{i=1}^{n_3}\|(\mathcal{Z}\times_3 \mathbf{U})^{(i)}\|_*=\|\mathcal{Z}\|_\text{TTNN}
\end{align*}
Denote the last element of set $Q$ be $q_L$ (suppose $Q$ is not empty), then we can rewrite and obtain the following formula:
\begin{equation*}
\mathcal{X}~\bar\star_{p\in P}\mathbf{U}_p \star_{q\in Q} \mathbf{U}_q =\mathcal{X}~\bar\star_{p\in P}\mathbf{U}_p\star_{q\in(Q/q_L)}\mathbf{U}_q \star_{q_L}\mathbf{U}_{q_L}
\end{equation*}
Let $\mathcal{P}:= \mathcal{X}~\bar\star_{p\in P}\mathbf{U}_p\star_{q\in(Q/q_L)}\mathbf{U}_q$ be a temporary tensor in which $/$ represents the set division, then we have 
\begin{equation*}
\|\mathcal{X}\|_\text{MNT-TNN}=\sum_{i=1}^{n_3}\|(\mathcal{P}\times_{q_L}\mathbf{U}_{q_L})\|_*
\end{equation*}
In the following, we consider two cases, i.e., $C(q_L)=1$ and $C(q_L)=2$.
First, when $C(q_L)=1$, it indicates that $\times_{q_L}$ represents mode-$\text{q}_L$ product, thus  
\begin{equation*}
\mathcal{P} \times_{q_L}\mathbf{U}_{q_L}=\text{Fold}_{q_L}(\mathbf{U}_{q_L}\mathbf{P}_{(q_L)})=\text{Fold}_{3}(\mathbf{U}_{q_L}Z_3)=\mathcal{Z}\times_3\mathbf{U}_{q_L}
\end{equation*}
where $\text{Unfold}(\mathcal{Z}, (1,2))=\text{Unfold}(\mathcal{P},(N/q_L))\in\mathbb{R}^{\times_{i\in N/q_L}n_i\times q_L}$. Hence $\mathcal{Z}$ is the desired agent tensor and $\mathbf{U}_{q_L}$ is the corresponding transformation factor; this case is done.

Second, when $C(q_L)=2$, w.l.o.g, we assume $q_L=\{k,p\}$ and take again $\mathcal{P}:= \mathcal{X}~\bar\star_{p\in P}\mathbf{U}_p\star_{q\in(Q/q_L)}\mathbf{U}_q$.

Since 
\begin{align*}
\mathcal{P}\star_{(k,p)}\mathbf{U}_{q_L}&=\text{Fold}_{(k,p)}(\mathbf{U}_{q_L}[\mathbf{P}^{(1)}_{[(k,p)]}~~\mathbf{P}^{(2)}_{[(k,p)]}\cdots~~\mathbf{P}^{(\frac{n_1n_2n_3}{n_kn_p})}_{[(k,p)]}])\\
&=\mathcal{P}\times_k\mathbf{U}_{q_L}\\
&=\text{Fold}_{N/k}(\mathbf{P}_{(N/k)})\times_3 \mathbf{U}_{q_L}
\end{align*}
thus clearly, $\mathcal{Z} = \text{Fold}_{N/k}(\mathbf{P}_{(N/k)})$ is the desired agent tensor. 

The analyses above demonstrate that the results of the two proposed mode products performed on the original tensor with orthogonal factors are equivalent to applying a mode-3 transform to a transformed surrogate tensor. Consequently, both cases reduce to the assertion in Lemma \ref{lemma:1}, thereby proving the statement of Theorem \ref{theorem}. 

In fact, even though the order between $\bar\star$ and $\star$ is reversed in Eq. (\ref{eq:transform}), the assertion remains valid; however, the roles of the transform factors and the original tensor must be exchanged. This conclusion can be obtained in a manner similar to the above discussions. In the following, we present two properties/effects of the two 2D mode products in problem Eq. (\ref{eq:prob}).

(1) The product denoted by $\bar\star$ helps introduce a latent spatiotemporal graph convolution operator.

\textit{proof.} We leave the proof of this property in Section \ref{sgtrans}.

(2) The effect of the product denoted as $\star$ can be shown by taking a special instance, that is, if $\mathbf{T}=\mathbf{I}$, we shall have $\|\frac{\mathcal{X}}{\gamma}\|_\text{MT-TNN}\le \alpha\|\mathcal{X}\|_\text{TTNN}$ where $\gamma$ is real and $\alpha$ is a positive number determined by $\gamma$ and $\|\mathbf{H}\|_*$.

\textit{proof.} Let us move out temporarily the transform leaded by $\mathbf{G}$, then we have $\mathcal{C}=\mathcal{Z}\times_{3}\mathbf{I}$ where $\mathcal{Z}=\mathcal{X}~\star_{(1,3)}\mathbf{H}$. Since $\mathcal{X}~\star_{(1,3)}\mathbf{H} = \mathcal{X}~\times_1 \mathbf{H}=\mathcal{X}~\star_{(1,2)}\mathbf{H}$ , then we have the following:
\begin{align*}
\|\frac{\mathcal{X}}{\gamma}\|_\text{MT-TNN}
&=\frac{1}{|\gamma|}\sum_{i=1}^{m_3}\|\mathbf{C}^{(i)}\|_*
=\frac{1}{|\gamma|}\sum_{i=1}^{m_3}\|\mathbf{Z}^{(i)}\|_*\\
&=\frac{1}{|\gamma|}\sum_{i=1}^{m_3}\|(\mathcal{X}\star_{(1,2)}\mathbf{H})^{(i)}\|_*\\
&=\frac{1}{|\gamma|}\sum_{i=1}^{m_3}\|\mathbf{HX}^{(i)}\|_*
\le \frac{1}{|\gamma|}\sum_{i=1}^{m_3}\|\mathbf{H}\|_*\|\mathbf{X}^{(i)}\|_*\\
&=\frac{\|\mathbf{H}\|_*}{|\gamma|}\sum_{i=1}^{m_3}\|(\mathcal{X}\times_3\mathbf{I})^{(i)}\|_*
=\frac{\|\mathbf{H}\|_*}{|\gamma|}\|\mathcal{X}\|_\text{TTNN}
\end{align*}

{A direct consequence of Theorem \ref{theorem} is that recovery theorems, such as Theorem \ref{the:2}, also apply to the extended multi-mode transform without nonlinear activation. While it is possible to extend this result to some decomposable nonlinear activations, such as kernel functions, by deriving and modifying the corresponding agent tensor, achieving a similar result for general nonlinearity remains challenging.}

\subsection{MNT-TNN Optimization}
As aforementioned, spatiotemporal traffic data discussed in this work consists of spatial, temporal, and joint spatiotemporal modalities, each potentially exhibiting distinct low-rank properties. In particular, our traffic tensor, denoted by $\mathcal{X}\in\mathbb{R}^{n_1\times n_2\times n_3}$, is a multivariate time series of 2D spatial graphs $\mathcal{G}$, where the first two dimensions describe jointly a scaled geographical location image, and the last dimension represents the time frames. Based on this observation, we formulate the following optimization problem:
\begin{gather}
\label{eq:prob}
\min_{\mathcal{X},\mathcal{C},\mathbf{G,H,T}}\sum_{i=1}^{m_3}\Vert{\psi(\mathbf{C})}^{(i)}\Vert_*\\ \nonumber
\mathit{s.t.}~~P_{\Omega}(\mathcal{X})=P_{\Omega}(\mathcal{O}),~
\mathcal{C}=\mathcal{X}~\bar{ \star}_{(1,2)}\mathbf{G}\star_{(1,3)} \mathbf{H}\times_3 \mathbf{T},\\ \nonumber \mathbf{G^\top G}=\mathbf{I}_{n_1n_2\times n_1n_2},\mathbf{H^\top H}=\mathbf{I}_{n_1\times n_1},\mathbf{T^\top T}=\mathbf{I}_{n_3\times n_3}.
\end{gather}
Here, $\mathcal{O}\in\mathbb{R}^{n_1\times n_2\times n_3}$ to be the tensor containing the observed data, and $\Omega$ to be the corresponding indicator tensor. $P_{\Omega}$ denotes the projection operator onto the tensor $\Omega$. Notably, the problem most closely related to ours has been defined in \cite{LiZJZH22}, which solely contains a 1D tube-wise transform. In contrast, we construct the problem from a more general and comprehensible perspective that aligns with the inherent characteristics of spatiotemporal tensors. Specifically, we have introduced two additional transforms, $\textbf{G}$ and $\textbf{H}$, to help extract the multimode low-rankness from the compact spatiotemporal tensor structure.
\subsection{Solving algorithm of MNT-TNN}
\label{subproblems}
In this part, we first introduce the optimization algorithm used for solving the proposed MNLRO problem (The terms, MNT-TNN and MNLRO, refer to the same concept in this paper). Then we present the derivations of the closed-form solutions. The proof for the convergence of this problem is placed at the end of this chapter.  

We first introduce an auxiliary variable $\mathcal{Z}=\phi(\mathcal{C})$ and turn Eq. (\ref{eq:prob}) into the following unconstrained problem using the half-quadratic splitting tricks \cite{krishnan2009fast}:
\begin{align}
\label{eq:10}
&L(\mathcal{X,C,Z},\mathbf{G,H,T})
:=\min_{\substack{\mathcal{X},\mathcal{C},\mathcal{Z},\\\mathbf{G},\mathbf{H},\mathbf{T}}}~\sum_{i=1}^{n_3}\Vert{\mathbf{Z}}^{(i)}\Vert_*\\& \nonumber
+\frac{\alpha}{2} \Vert\mathcal{C}-\mathcal{X} \bar{\star}_{(1,2)} \mathbf{G}\star_{(1,3)} \mathbf{H}\times_3 \mathbf{T}\Vert_F^2\\& \nonumber
+\frac{\beta}{2}\Vert \mathcal{Z}-\psi(\mathcal{C})\Vert_F^2 + \Phi(\mathcal{X})+\Upsilon{(\mathbf{G})}+\Upsilon{(\mathbf{H})}+\Upsilon{(\mathbf{T})}
\end{align}
where we introduce the following two indicator functions:
\begin{align*}
&\Phi(\mathcal{X})=
\begin{cases} 
0,  & P_{\Omega}(\mathcal{X})=P_{\Omega}(\mathcal{O}), \\
+\infty, & \text{otherwise}
\end{cases}
\\ \nonumber
&\Upsilon({\mathbf{U}})=
\begin{cases} 
0,  & \mathbf{U}^\top \mathbf{U}=I, \\
+\infty, & \text{otherwise}
\end{cases}
\end{align*}
and $\alpha$, $\beta> 0$ are penalty parameters. Then, we can alternatively upgrade each optimization variable in the following order:
\begin{equation}
\label{eq:process}
\begin{cases}
\mathcal{X}^{k+1} \in \mathop{\arg\min}\limits_{\mathcal{X}}\{L(\mathcal{X},\mathcal{C},\mathcal{Z},\mathbf{G,H,T}) + \frac{\rho_1}{2}\Vert\mathcal{X}-\mathcal{X}^{k}\Vert_F^2\} \\
\mathcal{Z}^{k+1} \in \mathop{\arg\min}\limits_{\mathcal{Z}}\{L(\mathcal{X},\mathcal{C},\mathcal{Z},\mathbf{G,H,T}) + \frac{\rho_2}{2}\Vert\mathcal{Z}-\mathcal{Z}^{k}\Vert_F^2\}\\
\mathcal{C}^{k+1} ~\in \mathop{\arg\min}\limits_{\mathcal{C}}\{L(\mathcal{X},\mathcal{C},\mathcal{Z},\mathbf{G,H,T}) + \frac{\rho_3}{2}\Vert\mathcal{C}-\mathcal{C}^{k}\Vert_F^2\}\\
\mathbf{G}^{k+1} \in \mathop{\arg\min}\limits_{\mathbf{G}}\{L(\mathcal{X},\mathcal{C},\mathcal{Z},\mathbf{G,H,T}) + \frac{\rho_4}{2}\Vert \mathbf{G-G}^{k}\Vert_F^2\}\\
\mathbf{H}^{k+1} \in \mathop{\arg\min}\limits_{\mathbf{H}}\{L(\mathcal{X},\mathcal{C},\mathcal{Z},\mathbf{G,H,T}) + \frac{\rho_5}{2}\Vert \mathbf{H-H}^{k}\Vert_F^2\}\\
\mathbf{T}^{k+1} \in \mathop{\arg\min}\limits_{\mathbf{T}}\{L(\mathcal{X},\mathcal{C},\mathcal{Z},\mathbf{G,H,T}) + \frac{\rho_6}{2}\Vert \mathbf{T-T}^{k}\Vert_F^2\}\\
\end{cases}
\end{equation}
Here we give derivations of the solutions of each subproblem.

 1. \textbf{Solving} $\mathcal{X}$ \textbf{subproblem}:
\begin{align}
\label{eq:x}
\nonumber
&\mathop{\arg\min}\limits_{\mathcal{X}}\{\frac{\alpha}{2} \Vert\mathcal{C}-\mathcal{X}\bar{ \star}_{(1,2)} \mathbf{G}\star_{(1,3)} \mathbf{H}\times_3 \mathbf{T}\Vert_F^2+\Phi(\mathcal{X})\\\nonumber&+\frac{\rho_1}{2}\Vert\mathcal{X}-\mathcal{X}^{k}\Vert_F^2\}\\
\nonumber
=& \mathop{\arg\min}\limits_{\mathcal{X}}\{\frac{\alpha}{2} \Vert\mathcal{C}\times_3 \mathbf{T}^\top\star_{(1,3)} \mathbf{H}^\top\bar{\star}_{(1,2)} \mathbf{G}^\top-\mathcal{X}\Vert_F^2\\\nonumber&+\frac{\rho_1}{2}\Vert\mathcal{X}-\mathcal{X}^{k}\Vert_F^2+\Phi(\mathcal{X})\}\\
\nonumber
=& \mathop{\arg\min}\limits_{\mathcal{X}}\{\frac{\alpha}{2} \Vert \mathcal{K}-\mathcal{X}\Vert_F^2+\frac{\rho_1}{2}\Vert\mathcal{X}-\mathcal{X}^{k}\Vert_F^2+\Phi(\mathcal{X})\}\\
\nonumber
=& \mathop{\arg\min}\limits_{\mathcal{X}}\{\frac{\alpha}{2} \Vert \mathbf{K}_{(3)}-\mathbf{X}_{(3)}\Vert_F^2+\frac{\rho_1}{2}\Vert\mathbf{X}_{(3)}-\mathbf{X}_{(3)}^{k}\Vert_F^2\}, \\\nonumber&~~s.t.~P_\Omega(\mathcal{X})=P_\Omega{(\mathcal{O})}\\ 
=&\left[(\alpha \mathcal{K}+\rho_1\mathcal{X}^k)/(\alpha+\rho_1)\right]_{\Omega^\top}+\mathcal{O}_\Omega
\end{align}
where $\mathcal{K}=\mathcal{C}\times_3 \mathbf{T}^\top\star_{(1,3)} \mathbf{H}^\top\bar{\star}_{(1,2)} \mathbf{G}^\top$, $\Omega^\top$ denotes the complement of set $\Omega$.

2. \textbf{Solving} $\mathcal{Z}$ \textbf{subproblem}:

Since
\begin{align}
\label{eq:z}
\nonumber
&\mathop{\arg\min}\limits_{\mathcal{Z}}\{\Vert\mathcal{Z}\Vert_*+\frac{\beta}{2}\Vert\mathcal{Z}-\psi(\mathcal{C})\Vert_F^2 + \frac{\rho_2}{2}\Vert \mathcal{Z}-\mathcal{Z}^k\Vert_F^2\}\\
\nonumber
=&\mathop{\arg\min}\limits_{\mathbf{Z}^{(i)}}\{\sum_{i=1}^{m_3}(\Vert\mathbf{Z}^{(i)}\Vert_*+\frac{\beta}{2}\Vert\mathbf{Z}^{(i)}-\psi(\mathbf{\mathbf{C}})^{(i)}\Vert_F^2 \\\nonumber&+ \frac{\rho_2}{2}\Vert \mathbf{Z}^{(i)}-(\mathbf{Z}^k)^{(i)}\Vert_F^2)\}\\
\nonumber
=&\mathop{\arg\min}\limits_{\mathbf{Z}^{(i)}}\{\sum_{i=1}^{m_3}(\Vert\mathbf{Z}^{(i)}\Vert_*+\frac{\beta+\rho_2}{2}\Vert\mathbf{Z}^{(i)}-({\beta\psi(\mathbf{C})}^{(i)}\\\nonumber&+\rho_2 (\mathbf{Z}^k)^{(i)}/(\beta+\rho_2))\Vert_F^2\}
\end{align}
It is equivalent to solving $m_3$ independent singular value thresholding problems, and the solution is
\begin{gather}
\Gamma_{\frac{1}{\beta+\rho_2}}[(\beta\psi(\mathbf{C}^{(i)})+\rho_2(\mathbf{Z}^{k})^{(i)})/(\beta+\rho_2)],~~i=1,2,...,m_3.
\end{gather}
where $\Gamma$ denotes the singular value shrinkage operator defined in \cite{cai2010singular}.

3. \textbf{Solving} $\mathcal{C}$ \textbf{subproblem}:
\begin{align}
\label{eq:c}
\nonumber
&\mathop{\arg\min}\limits_{\mathcal{C}}\{(\frac{\alpha}{2}\Vert\mathcal{C}-\mathcal{X}\bar{ \star}_{(1,2)} \mathbf{G}\star_{(1,3)} \mathbf{H}\times_3 \mathbf{T}\Vert_F^2 \\\nonumber&+ \frac{\beta}{2}\Vert \mathcal{Z}-\psi(\mathcal{C})\Vert_F^2 + \frac{\rho_3}{2}\Vert\mathcal{C}-\mathcal{C}^k\Vert_F^2\}\\
\nonumber
=&\mathop{\arg\min}\limits_{\mathcal{C}}\{\frac{\alpha}{2}\Vert\mathcal{C}-\mathcal{P}\Vert_F^2 + \frac{\beta}{2}\Vert \mathcal{Z}-\psi(\mathcal{C})\Vert_F^2 \}\\
\nonumber
=&\mathop{\arg\min}\limits_{C_{i,j,l}}\{\sum_{i,j,l}(\frac{\alpha+\rho_3}{2}(C_{i,j,l}-P_{i,j,l})^2\\&+\frac{\beta}{2}(Z_{i,j,l}-\psi(C_{i,j,l}))^2)\}
\end{align}
where $\mathcal{P}=(\alpha\mathcal{X}\bar{\star}_{(1,2)} \mathbf{G} \star_{(1,3)} \mathbf{H}\times_3 \mathbf{T}+\rho_3\mathcal{C}^k)/(\alpha+\rho_3)$. Note that this subproblem is decomposed into $m_1m_2m_3$ independent 1D minimization problems, and each of them can be solved efficiently by the Newton method.

4. \textbf{Solving} $\mathbf{G}$ \textbf{subproblem}:
\begin{align}
\label{eq:g}
&\mathop{\arg\min}\limits_{\mathbf{G}}\{\frac{\alpha}{2}\Vert\mathcal{C}-\mathcal{X}\bar{ \star}_{(1,2)} \mathbf{G}\star_{(1,3)} \mathbf{H}\times_3 \mathbf{T}\Vert_F^2 + \phi(G)
\nonumber \\&+ \frac{\rho_4}{2}\Vert \mathbf{G}-\mathbf{G}^k\Vert_F^2\}\\
\nonumber
=&\mathop{\arg\min}\limits_{\mathbf{G}}\{\frac{\alpha}{2}\Vert\mathcal{Y}-\mathcal{X}\bar{\star}_{(1,2)} \mathbf{G}\Vert_F^2  + \frac{\rho_4}{2}\Vert \mathbf{G}-\mathbf{G}^k\Vert_F^2+ \phi(\mathbf{G})\}\\
\nonumber
=&\mathop{\arg\min}\limits_{G}\{\frac{\alpha}{2}\Vert\mathbf{Y}_{(3)}-\mathbf{X}_{(3)} \mathbf{G}^\top\Vert_F^2  + \frac{\rho_4}{2}\Vert \mathbf{G}-\mathbf{G}^k\Vert_F^2+ \phi(\mathbf{G})\}\\
\nonumber
=&\mathop{\arg\min}\limits_{\mathbf{G},~\mathbf{G}^\top \mathbf{G}=\mathbf{I}}\{-\alpha \langle \mathbf{Y}_{(3)},\mathbf{X}_{(3)}\mathbf{G}^\top\rangle-\rho_4\langle \mathbf{G},\mathbf{G}^k\rangle\}\\
\nonumber
=&\mathop{\arg\max}\limits_{\mathbf{G},~\mathbf{G}^\top \mathbf{G}=\mathbf{I}}\{\mathbf{Tr}(\alpha \mathbf{Y}_{(3)}^\top \mathbf{X}_{(3)}\mathbf{G}^\top+\rho_4 \mathbf{G}^\top \mathbf{G}^k)\}\\
\nonumber
=&\mathop{\arg\max}\limits_{\mathbf{G},~\mathbf{G}^\top \mathbf{G}=I}\{\mathbf{Tr}\left[(\alpha \mathbf{X}_{(3)}^\top\mathbf{Y}_{(3)} +\rho_4(\mathbf{G}^k)^\top)\mathbf{G}\right]\}\\
\nonumber
\end{align}
where $\mathcal{Y}=\mathcal{C}\times_3 \mathbf{T}^\top\star_{(1,3)} \mathbf{H}^\top$, and we use the fact that $\mathcal{X}\bar{\star}_{(1,2)}\mathbf{G}=\mathbf{X}_{(3)}\mathbf{G}^\top$.
This problem is a variant of the orthogonal Procrustes problem \cite{schonemann1966generalized} and the unique optimal solution is $\mathbf{VU}^\top$, $\mathbf{U~\text{and}~V}$ come from the SVD of the matrix: $\alpha \mathbf{X}_{(3)}^\top\mathbf{Y}_{(3)} +\rho_4(\mathbf{G}^k)^\top\triangleq \mathbf{E}=\mathbf{U}\Sigma V^\top$.

The closed-form solutions of the remaining $\mathbf{H, T}$ subproblems can be derived in the same manner as $\mathbf{G}$'s, so we omit them due to the page limitation. The complete solution-finding process is enumerated in Algorithm 1. The convergence analysis is detailed in \ref{app:conv}.
\begin{algorithm}[tbh!]
    \caption{The PAM solving algorithm for MNT-TNN}
    \label{alg:1}
    \textbf{Input:} The observed tensor $\mathcal{O} \in \mathbb{R}^{n_1\times n_2\times n_3}$, the index set $\Omega$, transform factors $\mathbf{G, H, T}$, parameters $\rho_i,~i=1,2,\dots,6$, $\alpha~\text{and}~\beta.$ \\
    \textbf{Output:} The imputated result tensor $\mathcal{X}$.\\
    \textbf{Initialization:}$\mathcal{X}_0, \mathcal{C}_0, \mathcal{Z}_0, \mathbf{G}_0, \mathbf{H}_0, \mathbf{T}_0.$\\
    \begin{algorithmic}[1]
        \WHILE{$\frac{\Vert \mathcal{X}^{k+1} - \mathcal{X}^k\Vert_F}{\Vert \mathcal{X}^k \Vert_F} \le 10^{-4}$}
        \STATE Update $\mathcal{X}$ via Eq. (\ref{eq:x});\\
        \STATE Update $\mathcal{Z}$ via Eq. (\ref{eq:z});\\
        \STATE Update $\mathcal{C}$ via Eq. (\ref{eq:c});\\
        \STATE Update $\mathbf{G}$ via Eq. (\ref{eq:g});\\
        \STATE Update $\mathbf{H}$ via an equation analogous to Eq. (\ref{eq:g});\\
        \STATE Update $\mathbf{T}$ via an equation analogous to Eq. (\ref{eq:g});\\
    \ENDWHILE
    \end{algorithmic}
\end{algorithm}
\vskip -0.2in
% An interesting aspect of our MNT-TNN is the careful consideration of connections to existing works during its design, allowing for potential further explorations that build on the core ideas and motivations behind our model.
\vskip -0.2in
\subsection{Interrelationship with HOSVD}
Recall that the TTNN optimization essentially extracts the low-rankness of the frontal slices of a tensor in a certain transform domain; however, the tensor structure is orientation dependent, which means that applying transforms to different modes shall lead to different results. However, our model described in Eq. (\ref{eq:10}) takes all modes and slices into account. As can be seen from Eq. (\ref{eq:c}) and Eq. (\ref{eq:g}), the shared formula $\Vert\mathcal{C}-\mathcal{X}\bar{ \star}_{(1,2)} \mathbf{G}\star_{(1,3)} \mathbf{H}\times_3 \mathbf{T}\Vert_F^2$ can be also written as $\Vert\mathcal{X}-\mathcal{C}\times_3 \mathbf{T}^\prime\star_{(1,3)} \mathbf{H}^\prime\bar{ \star}_{(1,2)} \mathbf{G}^\prime\Vert_F^2$ where $\mathbf{T}^\prime$, $\mathbf{H}^\prime$, and $\mathbf{G}^\prime$ are auxiliary variables introduced to represent the transpose of these factor matrices. This actually demonstrates that these subproblems are partially in agreement with the standard HOSVD/Tucker factorization problem,
\begin{align}
\min\limits_{\substack{\mathcal{X},\mathcal{C}, \\\mathbf{U}_1,\mathbf{U}_2,\mathbf{U}_3}}&\frac12 \|\mathcal{X}-C\times_1 \mathbf{U}_1\times_2\mathbf{U}_2\times_3\mathbf{U}_3\|_F^2\\\nonumber s.t.& ~~\mathbf{U}_i^\top \mathbf{U}_i=\mathbf{I},~~rank(\mathbf{U}_i)\le r_i
\end{align}
and indeed can be seen as a pair of dual generalized Tucker decomposition proximal operators with certain prior knowledge. Furthermore, this observation indicates that MNLRO can inherently incorporate auxiliary information encoded in the factor priors of tensors, as discussed in \cite{chen2013tensorfactor}. Intuitively, as will be demonstrated in the experimental section, the MNLRO model fundamentally seeks a latent low-rank kernel within a composed transformed domain, aided by specific linear transforms and nonlinear activations performed independently. According to the discussions in \cite{goulart2017traffictuc}, the resulting core $\mathcal{C}$ is conceivably compressible without necessarily being sparse.

\subsection{Spatiotemporal Graph Transformation of MNT-TNN}
\label{sgtrans}
Employing our definitions in Section \ref{method}, we argue that there is a latent spatiotemporal graph transformation within the generalized Tucker decomposition constraint in Eq. (\ref{eq:10}).

\textit{proof}
Let us move out temporally the transformation leaded by $\mathbf{H}$, i.e., we have $\mathcal{C}=\mathcal{X} \bar{ \star}_{(1,2)} \mathbf{G}\times_3 \mathbf{T}$.
Since $\mathcal{X}\bar{ \star}_{(1,2)} \mathbf{G}=\mathbf{Fold}_{(1,2)}(\mathbf{G}\overline{\mathbf{X}}_{[(1,2)]})=\mathbf{Fold}_{(3)}[(\mathbf{G}{\mathbf{X}}_{(3)}^\top)^\top]=\mathbf{Fold}_{(3)}(\mathbf{X}_{(3)} \mathbf{G}^\top)$, 
thus $\mathcal{C}=\mathcal{X} \bar{ \star}_{(1,2)} \mathbf{G}\times_3 \mathbf{T}=\mathbf{Fold}_{(3)}(\mathbf{T}\mathbf{X}_{(3)} \mathbf{G}^\top)\implies \overline{\mathbf{C}}_{(1,2,3)}=(\mathbf{T}\otimes \mathbf{G})\overline{\mathbf{X}}_{(1,2,3)}.$ 
where $\otimes$ denotes the Kronecker product and the formula $(\mathbf{T}\otimes \mathbf{G})\overline{\mathbf{X}}_{(1,2,3)}$ is exactly a separable spatio-temporal filtering operation defined by \cite{pan2021spatiotemporal}, because of this, the spatiotemporal traffic tensor can enjoy the spatial, temporal and spatiotemporal exploration in our method, simultaneously.

\subsection{Augmented TTNN families optimizer (ATTNNs)}
\begin{figure*}[hbtp]
% \vskip 0.2in
\centering
\begin{minipage}[hbtp]{\linewidth}
\includegraphics[width=\linewidth]{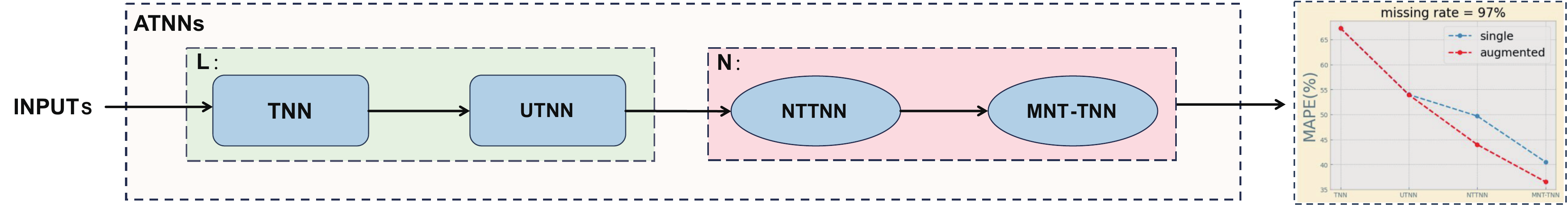}
\end{minipage}
\caption{The model of ATNNs used in this paper, in which every optimizer operates on the tensor sequentially. The rightmost line chart displays the numerical results in terms of MAPE of each optimizer within the ATNNs, in comparison to the performance of each optimizer alone.}
\label{fig:3}
% \vskip -0.2in
\end{figure*}
Within the framework of TTNN, existing TTNN methods (hereafter referred to as optimizers) can individually achieve satisfactory performances in tensor recovery tasks. Thus, it is a promising endeavor to further enhance the recovery performance of these methods by combining them to leverage their respective strengths. Specifically, according to the recovery theorems such as Theorem \ref{the:2}, the performance of these \textit{linear} optimizers is prone to meet sharp declines under high missing rates in practice. On the contrary, \textit{nonlinear} optimizers are highly dependent on the quality of initial input, whereas linear ones, being convex, are less affected by initialization qualities. Therefore, it is reasonable to exploit such a compensation relationship using boosting skills. As shown in Fig. \ref{fig:3}, a simple consecutive structure can significantly improve the imputation performance at high missing ratios.

Technically, we categorize the four selected optimizers into two groups, a linear group comprising $\mathbf{L}:\text{TNN}\rightarrow \text{UTNN}$ and a nonlinear group comprising $\mathbf{N}:\text{NTTNN} \rightarrow \text{MNT-TNN}$. The internal order within each group is based on the optimizers' capabilities. By Theorem \ref{the:2}, even though the capabilities of linear models are restricted due to their reliance upon the sampling ratio, we expect that the linear group can provide a reliable and relatively strong initial estimate for the nonlinear group. This initial estimate, when refined by the nonlinear group, allows the nonlinear optimizers to overcome prior limitations, producing results superior to those achieved by any individual optimizer.

\section{Experiments}
\label{experiment}
\begin{figure}[htbp]
% \vskip 0.2in
\centering
\includegraphics[scale=0.3]{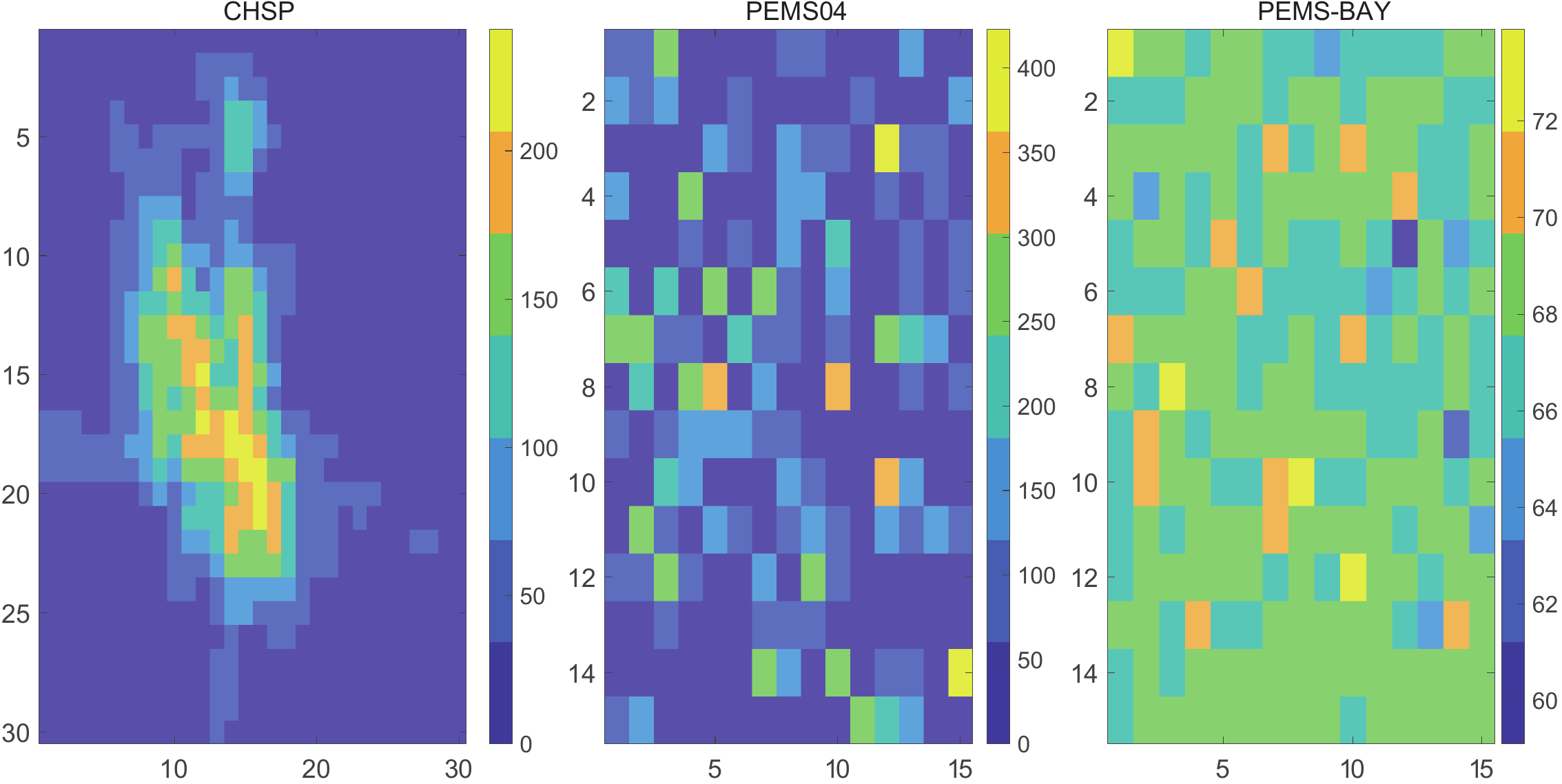}
\caption{Visualization of frontal slices of tensors taken from the three used datasets. Each dataset shows a distinct spatial pattern, and that of CHSP implies the most obvious spatial dependency.}
\label{dataset}
% \vskip -0.2in
\end{figure}
\subsection{Datasets Descriptions}
Given the assumption of multimode low-rankness, the spatial dynamics within the compact spatio-temporal tensor may significantly influence the efficacy of MNT-TNN. Hence, to evaluate the effectiveness of our proposed methods, we carefully select three different real-world datasets tailored to our requirements, 
\begin{itemize}
    \item \textbf{CHSP}: Real-time car-hailing data collected via GNSS, comprising 475,200 raw data points organized into a 30$\times$30$\times$528 3D tensor spanning latitude, longitude, and time.
    \item \textbf{PEMS04}: A widely-used public traffic dataset for spatiotemporal forecasting and other relevant tasks \cite{liu2022scinet}. This dataset contains traffic network data from California collected every 5-minute across 307 sensors. For our experiment, we utilize data from the first 4 days and the first 225 sensors, reshaping it into a 15$\times$15$\times$1152 3D tensor where each element represents the traffic flow at a specific position and time. 
    \item \textbf{PEMS-BAY}: Public traffic data from California collected every 5-minute by 325 sensors \cite{liu2022scinet}. We choose the first 7 days from the first 225 sensors, forming a 15$\times$15$\times$2016 3D tensor where each element represents the traffic speed at a specific position and time.
\end{itemize}
As presented in Fig. \ref{dataset}, the frontal slices of these 3D tensor data explicitly reveal varying degrees of spatial dependencies. We can see that CHSP exhibits a dense geographical pattern with concentrated energy, while the spatial correlations in the other two datasets diffuse differently. 

We compare our proposed methods with other transform-based TNN methods, including $\text{TNN}$ \cite{LuFCLLY20}, $\text{UTNN}$ \cite{SongNZ20}, $\text{NTTNN}$ \cite{LiZJZH22}, $\text{FTNN}$ \cite{JiangNZH20}, and LRTC models including HaLRTC \cite{liu2012lrtc} and LRTC-TNN \cite{chen2020nonconvex}. 
All these methods are introduced in section \ref{works}, with hyperparameters set according to the authors’ recommendations and further tuned for optimal performance. Additional hyperparameter settings for $\text{MNT-TNN}$ are detailed in the case study section.

In particular, with regard to the two nonlinear methods, we adopt the same strategy as suggested in \cite{LiZJZH22}, where a simple linear interpolation \cite{yair2018multi} is used to acquire an initial input, and the tangent hyperbolic function is deployed as the nonlinear activation. Considering that the lengths of the time dimensions of two PEMS datasets are relatively long, FTNN \cite{JiangNZH20} is more competitive due to its capability of frame feature capturing. So we take the output of FTNN as the initial input for nonlinear optimizers in ATTNNs when conducting experiments on the PEMS datasets. What's more, we adopt the early stopping strategy to record the best score of two nonlinear optimization problems since they are non-convex. 

All the experiments are conducted under Windows 10 and MATLAB R2019a running on a Laptop (AMD Ryzen 7 5800H, 3.20 GHz, 32GB RAM).
\subsection{Settings of Transform Matrices}
\label{subsec:settings of factors}
\begin{algorithm}[tbh!]
    \caption{{Pipeline for obtaining the transform factors}}
    \label{alg:TransFac}
    \textbf{Input:} The node set $\mathbf{V}$ with size $\vert \mathbf{V}\vert= n^2$, the distance $d_{ij}~,i,j=1,2,...,n^2$ between each two nodes in $\mathbf{V}$.\\
    \textbf{Output:} The transform factors $\mathbf{G}, \mathbf{H}$. \\
    % \textbf{Initialization:}$\mathcal{X}_0, \mathcal{C}_0, \mathcal{Z}_0, \mathbf{G}_0, \mathbf{H}_0, \mathbf{T}_0.$\\
    \noindent
    \begin{algorithmic}[2]
          \STATE Construct the sparse adjacency matrix $\mathbf{A}\in \mathbb{R}^{n^2\rtimes n^2}$ through thresholded Gaussian kernel weighting function\cite{shuman2013emerging} and the distance set $[d_{ij}]$;\\
          \STATE Compute the degree matrix $\mathbf{D}$ according to $\mathbf{A}$ and the normalized graph laplacian by $\mathbf{\tilde A}=\mathbf{I}-\mathbf{D}^{-1/2}\mathbf{AD}^{-1/2}$;\\
          \STATE Derive the transform factor $\mathbf{G}$ from the the SVD of $\mathbf{\tilde A}$; \\
          \FOR{$k=1$ to $n$}
          \STATE Obatin $h_k\in\mathbb{R}^{n^2}$ by averaging along the column of the submatrix $[a_{:,~:kn}]$;\\
          \ENDFOR
          \STATE Construct the adjacency matrix $\hat{\mathbf{H}}\in \mathbb{R}^{n\times n}$ with $h_{ij}=h_i^\top h_j$;\\
          \STATE Compute the degree matrix $\hat{\mathbf{D}}$ according to  $\hat{\mathbf{H}}$ and the lazy random walk matrix $\Theta=\frac{1}{2}(\mathbf{I}+\mathbf{\hat D}^{-1}\mathbf{\hat H})$;\\
          \STATE Compute the scattering wavelets \cite{pan2021spatiotemporal}  by $\mathbf{H}_\Theta=\sum_{j=1}^JH_j(\Theta)=\sum_{j=1}^J\Theta^{2^j-1}(\mathbf{I}-\Theta^{2^j-1})$ ;\\
          \STATE Derive the transform factor $\mathbf{H}$ from the the SVD of $\mathbf{H}_\Theta$;\\
    \end{algorithmic}
\end{algorithm}
Selecting proper transform factors for low-rank exploration has consistently been a key task for TTNN algorithms. As the third mode transform, denoted by $\mathbf{T}$, has been extensively investigated in prior research, this section focuses on the factor matrices for the other two additional modes:$\mathbf{G}$ for the spatial mode and $\mathbf{H}$ for the spatiotemporal mode. 

Consider that the spatial modal is described by the frontal slices of the tensor, which are all $n$ by $n$ matrices carrying intrinsically the location information of all the positioning/sensing nodes, we can use the Hodge 0-laplacian \cite{lim2020hodge} also known as the graph laplacian to convolve the features of all nodes with underlying graph structure. Concretely, start by $n^2$ locations in the node set $\mathbf{V}$, we construct the adjacency matrix $\mathbf{A}\in\mathbb{R}^{n^2\times n^2 }$ using the gaussian kernel thresholding \cite{shuman2013emerging}. Then we can obtain the 1-step normalized graph laplacian $\mathbf{\tilde A}$. Lastly, the left singular vector matrix of $\mathbf{\tilde A}$ is used as the desired operator $\mathbf{G}$. It is known that this factor consists of the basic spectral components of the graph; hence by applying it to the folded vectors, we incorporate the spectral information into the transformation process and then reconstruct the matrix that reflects the graph structure and properties. (i.e., the operation $\bar\star_{(1,2)}$)

On the other hand, we can derive $\mathbf{H}$ from the established graph matrix trivially. Specifically, given the connections between each pair of nodes $\mathbf{A}\in \mathbb{R}^{n^2\times n^2}$, we can aggregate the information of nodes in the same longitude (rows in the scaled graph) by averaging each consecutive submatrix of size $n^2\times n$. Each results in a feature vector $h_i\in \mathbb{R}^{n^2},i=1,2\dots,n$ which roughly represents the geometric feature of one specific longitude. We then compute the adjacency matrix $\mathbf{\hat H}$ for longitudes by calculating the inner product $h_i^\top h_j,~i,j=1,2\dots,n$. Moreover, to enable a multiresolution analysis, we employ geometric scattering wavelets to filter the graph signal, as implemented by \cite{pan2021spatiotemporal}. Thus, we derive the singular vector matrix $\mathbf{H}$ of the matrix $\mathbf{H}_\Theta=\sum_{j=1}^JH_j(\Theta)=\sum_{j=1}^J\Theta^{2^j-1}(\mathbf{I}-\Theta^{2^j-1})$, where $\Theta=\frac{1}{2}(\mathbf{I}+\mathbf{\hat D}^{-1}\mathbf{\hat H})$ is the lazy random walk matrix and $\mathbf{\hat D}$ is the degree matrix of $\mathbf{\hat H}$.

Notably, these two transform factors are entirely determined by the geometric properties of the collected data and are independent of the data values.
\subsection{Comparison Results of MNT-TNN}
\label{compare}
\begin{table}[htbp]
\vskip -0.2in
\caption{Imputation accuracy on CHSP with normal missing rates.}
\label{tab:1}
% \begin{center}
% \begin{sc}
\resizebox{1.0\linewidth}{!}{
\begin{tabular}{c c c c c c c c c}
\toprule
\multirow{2}{*}{Methods} & \multicolumn{2}{c}{MR=90\%} & \multicolumn{2}{c}{MR=70\%} & \multicolumn{2}{c}{MR=50\%} & \multicolumn{2}{c}{MR=30\%} \\
\cmidrule(r){2-3} \cmidrule(r){4-5} \cmidrule(r){6-7} \cmidrule(r){8-9} 
 & MAPE(\%) & RMSE & MAPE(\%) & RMSE & MAPE(\%) & RMSE & MAPE(\%) & RMSE \\
\midrule
HaLRTC  & 56.40$\pm$0.22 & 33.81$\pm$0.19 & 33.38$\pm$0.06 & 12.25$\pm$0.12 & 25.70$\pm$0.12 & 4.63$\pm$0.04 & 23.24$\pm$0.23 & 3.60$\pm$0.03\\

LRTC-TNN & 58.63$\pm$0.65 & 11.59$\pm$0.16 & 32.96$\pm$0.17 & 5.67$\pm$0.05 & 28.39$\pm$0.09 & 4.07$\pm$0.005 & 25.11$\pm$0.05 & 3.50$\pm$0.02 \\
\midrule
TNN & 54.38$\pm$0.41 & 7.97$\pm$0.02 & 44.77$\pm$0.09 & 5.90$\pm$0.04 & 38.50$\pm$0.19 & 4.81$\pm$0.03 & 33.43$\pm$0.08 & 4.01$\pm$0.04\\

UTNN & 44.17$\pm$0.18 & 7.01$\pm$0.01 & 34.99$\pm$0.14 & 4.99$\pm$0.04 & 29.62$\pm$0.13 & 4.02$\pm$0.03 & 25.66$\pm$0.15 & 3.40$\pm$0.02\\

FTNN & 73.34$\pm$0.69 & 10.90$\pm$0.09 & 41.95$\pm$0.27 & 6.23$\pm$0.02 & 33.48$\pm$0.45 & 4.83$\pm$0.02 & 28.50$\pm$0.33 & 3.51$\pm$0.03\\

NTTNN & 36.93$\pm$0.17 & 6.48$\pm$0.03 & 30.27$\pm$0.10 & 4.71$\pm$0.05 & 25.95$\pm$0.10 & 3.83$\pm$0.01 & 24.33$\pm$0.13 & 3.46 $\pm$ 0.02\\
\midrule 

MNT-TNN(ours) & \underline{33.51}$\pm$0.20 & \underline{6.01}$\pm$0.07 &
\underline{28.59}$\pm$0.06 & \pmb{4.47}$\pm$0.04 & \underline{25.51}$\pm$0.09 & \underline{3.75}$\pm$0.02 & \underline{23.22}$\pm$0.06 & \underline{3.25}$\pm$0.03\\

ATTNNs(ours)  & \pmb{33.03}$\pm$0.14 & \pmb{5.87}$\pm$0.02 & \pmb{27.76}$\pm$0.09 & \underline{4.53}$\pm$0.04 & \pmb{24.67}$\pm$0.09 & \pmb{3.73}$\pm$0.02 & \pmb{22.26}$\pm$0.05 & \pmb{3.18}$\pm$0.04\\
\bottomrule
\end{tabular}
}
% \end{sc}
\vskip -0.05in
% \end{center}
\end{table}
\begin{table}[htbp]
\caption{Imputation accuracy on CHSP with high missing rates.}
\label{tab:2}
% \begin{center}
\resizebox{1.0\linewidth}{!}{
\begin{tabular}{c c c c c c c}
\toprule
\multirow{3}{*}{Methods} & \multicolumn{2}{c}{MR=97\%} & \multicolumn{2}{c}{MR=95\%} & \multicolumn{2}{c}{MR=93\%} \\
\cmidrule(r){2-3} \cmidrule(r){4-5} \cmidrule(r){6-7} 
 & MAPE(\%) & RMSE & MAPE(\%) & RMSE & MAPE(\%) & RMSE \\
\midrule
HaLRTC  & 74.82$\pm$0.42 & 46.97$\pm$0.24 & 66.74$\pm$0.17 & 42.77$\pm$0.25 & 65.67$\pm$2.24 & 41.81$\pm$2.05\\

LRTC-TNN & 91.15$\pm$0.57 & 26.10$\pm$0.15 & 81.88$\pm$0.15 & 19.72$\pm$0.25 & 71.00$\pm$0.48 & 15.61$\pm$0.23\\
\midrule
TNN & 67.27$\pm$0.36 & 10.46$\pm$0.26 & 60.06$\pm$0.52 & 9.18$\pm$0.13 & 59.40$\pm$1.69 & 9.05$\pm$0.34\\

UTNN & 53.99$\pm$0.26 & 10.13$\pm$0.36 & 48.44$\pm$0.31 & 8.28$\pm$0.12 & 47.91$\pm$1.24 & 8.12$\pm$0.39\\

FTNN & 159.79$\pm$1.78 & 25.51$\pm$0.47 & 113.89$\pm$0.83 & 17.72$\pm$0.34 & 108.44$\pm$11.45 & 16.79$\pm$2.00\\

NTTNN & 49.70$\pm$0.14 & \underline{8.45}$\pm$0.21 & 43.50$\pm$0.21 & \underline{7.46}$\pm$0.11 & 40.28$\pm$0.16 & 6.89$\pm$0.04\\
\midrule 

MNT-TNN(ours) & \underline{40.51}$\pm$1.01 & 9.92$\pm$0.76 &
\underline{37.17}$\pm$0.40 & 7.94$\pm$0.22 & \underline{35.00}$\pm$0.27 & \underline{6.69}$\pm$0.09\\

ATTNNs(ours)  & \pmb{36.53}$\pm$0.18 & \pmb{7.74}$\pm$0.12 &
\pmb{35.13}$\pm$0.14 & \pmb{6.99}$\pm$0.06
& \pmb{34.33}$\pm$0.16 & \pmb{6.48}$\pm$0.03\\
\bottomrule
\end{tabular}
}
% \end{center}
% \vskip -0.1in
\end{table}
We exclusively evaluate all methods across a normal range of missing rates (MRs) $\{30\%, 50\%, 70\%, 90\%\}$, as well as particularly high MRs $\{93\%, 95\%, 97\%\}$. Overall, the two proposed methods, MNT-TNN, and ATTNNs, achieve the best performance interchangeably among all compared methods. The main comparison results at standard MRs are clearly shown in Table. \ref{tab:1}, from which we observe that TTNN-based methods consistently outperform LRTC methods since they are more adapted for random missing patterns, and \text{MNT-TNN} can consistently outperform other TTNN-based methods. Nevertheless, we note that at low MRs, LRTC models are still competitive, and the improvements achieved by our augmented method, ATTNNs, are not yet considerable. As the MR gradually increases, MNT-TNN demonstrates a significant advantage in imputation performance. As shown in Table. \ref{tab:2}, the performances of LRTC models and \text{FTNN} nearly collapse, while \text{NTTNN} and \text{MNT-TNN} consistently outperform others. This trend reflects the limited recovery capacity of linear optimizers at high MRs. Concretely, compared with the second-best method, NTTNN, MNT-TNN achieves an average improvement of $12.2\%$ in terms of MAPE. In the meanwhile, the effect of ATTNNs also becomes transparent, delivering a $4.4\%$ boost over MNT-TNN.
\begin{figure}[htbp]
% \vskip 0.2in
\centering
\includegraphics[scale=0.6]{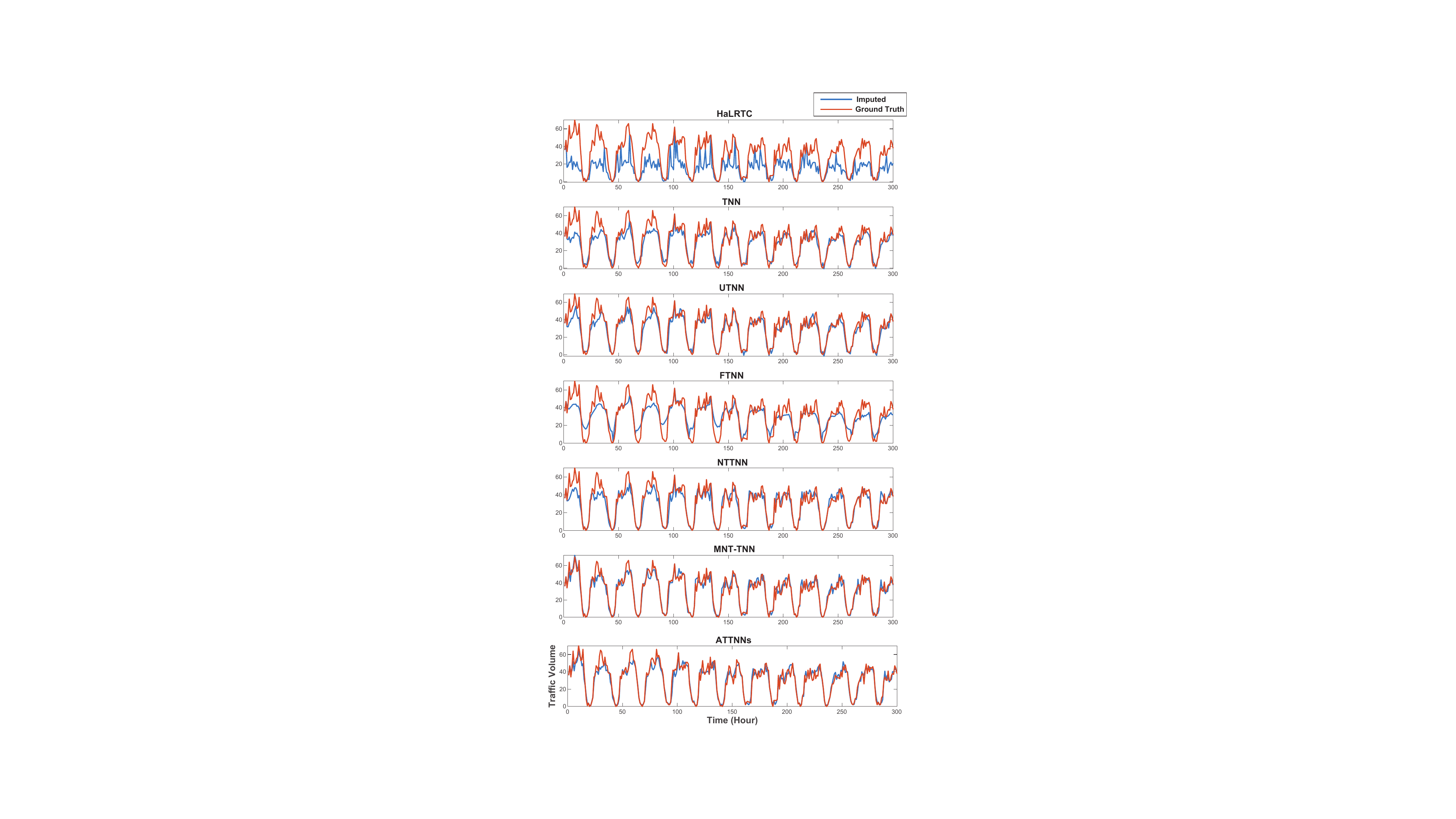}
\caption{Temporal view of the imputation results of compared methods with the $\text{MR}=90\%$.}
% \vskip -0.2in
\end{figure}

\begin{figure}[htbp]
% \vskip 0.2in
\centering
\begin{minipage}[htbb]{\linewidth}
\includegraphics[width=\linewidth]{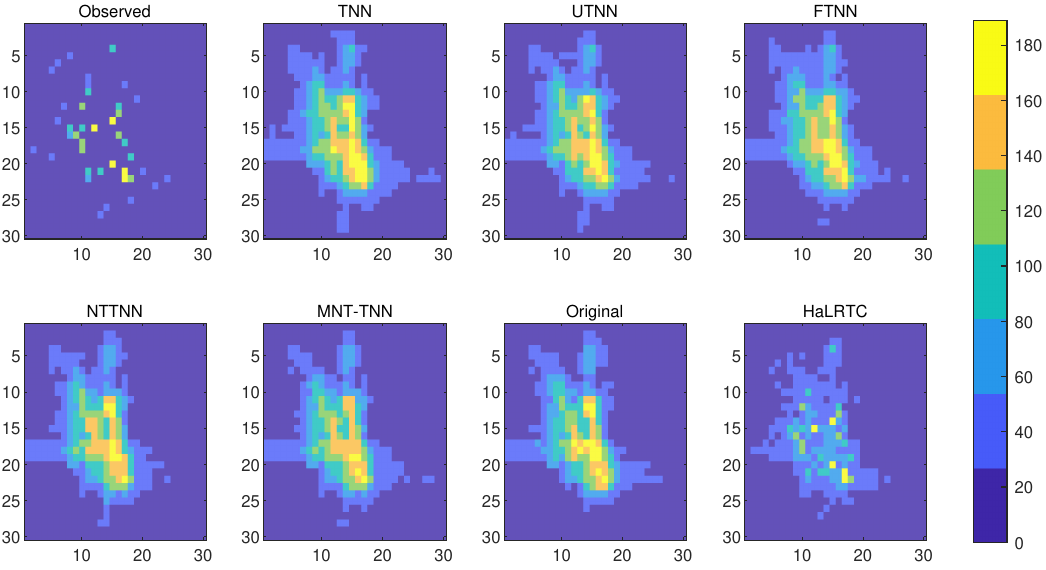}
\end{minipage}
\caption{Spatial view of the imputation results of compared methods with the $\text{MR}=90\%$.}
% \vskip -0.2in
\end{figure}

\begin{table}[htbp]
\caption{Imputation accuracy on PEMS04 with different missing rates.}
\label{tab:3}
% \begin{sc}
\resizebox{1.0\linewidth}{!}{
\begin{tabular}{c c c c c c c c c}
\toprule
\multirow{3}{*}{Methods} & \multicolumn{2}{c}{MR=90\%} & \multicolumn{2}{c}{MR=70\%} & \multicolumn{2}{c}{MR=50\%} & \multicolumn{2}{c}{MR=30\%} \\
\cmidrule(r){2-3} \cmidrule(r){4-5} \cmidrule(r){6-7} \cmidrule(r){8-9}
 & MAPE(\%) & RMSE & MAPE(\%) & RMSE & MAPE(\%) & RMSE & MAPE(\%) & RMSE \\
\midrule
HaLRTC  & 66.53 & 198.56 & 40.15 & 122.35 & 26.59 & 74.37 & 19.20 & 44.14\\

LRTC-TNN & 63.94 & 87.72 & 35.97 & 54.11 & 21.75 & 38.87 & 18.66 & 32.36\\
\midrule
TNN & 34.17 & 41.97 & 25.37 & 33.08 & 23.34 & 30.82 & 22.57 & 29.80\\

UTNN & 26.31 & 37.53 & 21.21 & 31.52 & 21.92 & 30.42 & 21.38 & 29.52\\

FTNN & \underline{18.31} & 34.21 & \underline{13.56} & 28.16 & \underline{12.85} & \underline{26.82} & \underline{12.66} & \underline{26.14}\\

NTTNN(FTNN) & 19.61 & 33.48 & 14.36 & \underline{27.82} & 13.84 & 27.06 & 13.72 & 26.81 \\
\midrule 

MNT-TNN(FTNN) & \pmb{17.97} & \pmb{33.27} &
\pmb{13.22} & \pmb{27.76} & \pmb{12.37} & \pmb{26.33} & \pmb{12.11} & \pmb{25.61}\\
\bottomrule
\end{tabular}
}

\resizebox{1.0\linewidth}{!}{
\begin{tabular}{c c c c c c c}
\toprule
\multirow{3}{*}{Methods} & \multicolumn{2}{c}{MR=97\%} & \multicolumn{2}{c}{MR=95\%} & \multicolumn{2}{c}{MR=93\%} \\
\cmidrule(r){2-3} \cmidrule(r){4-5} \cmidrule(r){6-7} 
 & MAPE(\%) & RMSE & MAPE(\%) & RMSE & MAPE(\%) & RMSE \\
\midrule

HALRTC  & 88.76 & 242.42 & 80.47 & 228.26 & 73.98 & 215.59\\

LRTC-TNN & 62.10 & 169.56 & 58.55 & 146.21 & 57.15 & 131.83\\
\midrule
TNN & 53.84 & 65.57 & 42.56 & 51.58 & 37.42 & 45.88\\

UTNN & 49.12 & 65.08 & 33.41 & 47.52 & 28.70 & 40.76\\

FTNN & \underline{33.65} & 53.61 & \underline{24.97} & 44.43 & \underline{21.39} & 39.17\\

NTTNN(FTNN) & 33.82 & \underline{50.56} & 25.70 & \pmb{41.86} & 21.40 & \pmb{36.79}\\
\midrule 
MNT-TNN(FTNN) & \pmb{32.53} & \pmb{50.07} &
\pmb{24.46} & \underline{41.99} & \pmb{20.55} & \underline{37.30}\\

\bottomrule
\end{tabular}
}
\end{table}

\begin{table}[htbp]
\centering
\caption{Imputation accuracy on PEMS-BAY with different missing rates.}
\label{tab:4}
% \begin{sc}
\resizebox{\textwidth}{!}{
\begin{tabular}{c c c c c c c c c}
\toprule
\multirow{3}{*}{Methods} & \multicolumn{2}{c}{MR=90\%} & \multicolumn{2}{c}{MR=70\%} & \multicolumn{2}{c}{MR=50\%} & \multicolumn{2}{c}{MR=30\%} \\
\cmidrule(r){2-3} \cmidrule(r){4-5} \cmidrule(r){6-7} \cmidrule(r){8-9}
 & MAPE(\%) & RMSE & MAPE(\%) & RMSE & MAPE(\%) & RMSE & MAPE(\%) & RMSE \\
\midrule
HaLRTC  & 82.89 & 52.56 & 67.03 & 42.95 & 50.78 & 32.70 & 28.79 & 18.30 \\

LRTC-TNN & 9.37 & 6.83 & 6.85 & 5.09 & 4.69 & 3.77 & 3.43 & 2.81\\
\midrule
TNN & 6.74 & 4.63 & 3.93 & 2.87 & 2.87 & 2.14 & 2.33 & 1.76\\

UTNN & 5.80 & 4.16 & 3.72 & 2.76 & 2.81 & 2.12 & 2.28 & 1.74\\

FTNN & \underline{4.86} & \underline{3.97} & \underline{2.63} & \underline{2.31} & \underline{2.02} & \underline{1.77} & \underline{1.70} & \underline{1.47}\\

NTTNN(FTNN) & 5.45 & 4.20 & 3.62 & 2.89 & 3.37 & 2.70 & 3.31 & 2.65\\
\midrule 

MNT-TNN(FTNN) & \pmb{4.82} & \pmb{3.91} &
\pmb{2.60} & \pmb{2.28} & \pmb{1.96} & \pmb{1.71} & \pmb{1.63} & \pmb{1.41}\\

\bottomrule
\end{tabular}
}
% \end{sc}
% \begin{sc}
\resizebox{\textwidth}{!}{
\begin{tabular}{c c c c c c c}
\toprule
\multirow{3}{*}{Methods} & \multicolumn{2}{c}{MR=97\%} & \multicolumn{2}{c}{MR=95\%} & \multicolumn{2}{c}{MR=93\%} \\
\cmidrule(r){2-3} \cmidrule(r){4-5} \cmidrule(r){6-7} 
 & MAPE(\%) & RMSE & MAPE(\%) & RMSE & MAPE(\%) & RMSE \\
\midrule
HaLRTC  & 90.32 & 56.94 & 87.88 & 55.51 & 85.76 & 54.25\\

LRTC-TNN & 14.64 & 10.95 & 10.28 & 7.50 & 9.78 & 7.14\\
\midrule
TNN & 10.62 & 6.93 & 8.88 & 5.91 & 7.77 & 5.25\\

UTNN & 10.29 & 6.84 & 8.22 & 5.60 & 6.87 & 4.81\\

FTNN & 9.13 & 6.81 & \underline{7.12} & 5.50 & \underline{5.95} & \underline{4.72}\\

NTTNN(FTNN) & \underline{8.98} & \underline{6.54} & 7.20 & \underline{5.35} & 6.27 & 4.76\\
\midrule 

MNT-TNN(FTNN) & \pmb{8.85} & \pmb{6.40} &
\pmb{7.06} & \pmb{5.32} & \pmb{5.88} & \pmb{4.62}\\
\bottomrule
\end{tabular}
}
% \end{sc}
% \vskip -0.1in
\end{table}
Beyond the CHSP dataset, which offers admirable spatial patterns for exploitation, we also evaluate imputation performance on the PEMS04 and PEMS-BAY datasets. Here, low-rank exploration demonstrates limited benefits from spatial mode transforms, leading to a diminished advantage for MNT-TNN. By contrast, FTNN achieves excellent imputation accuracy due to its frame-specific characteristics. Adopting the ATTNNs approach, we use the output of FTNN as input for two nonlinear methods, NTTNN and MNT-TNN. Table. \ref{tab:3} and Table. \ref{tab:4} show that MNT-TNN successfully enhances the imputation accuracy of FTNN, whereas NTTNN struggles to leverage this initialization. This finding not only validates the effectiveness of ATTNNs but also underscores the superiority of MNT-TNN over NTTNN {in terms of its capacity to manage diverse initial inputs}.
\begin{figure}[htbp]
% \vskip -0.2in
\centering
\includegraphics[scale=0.85]{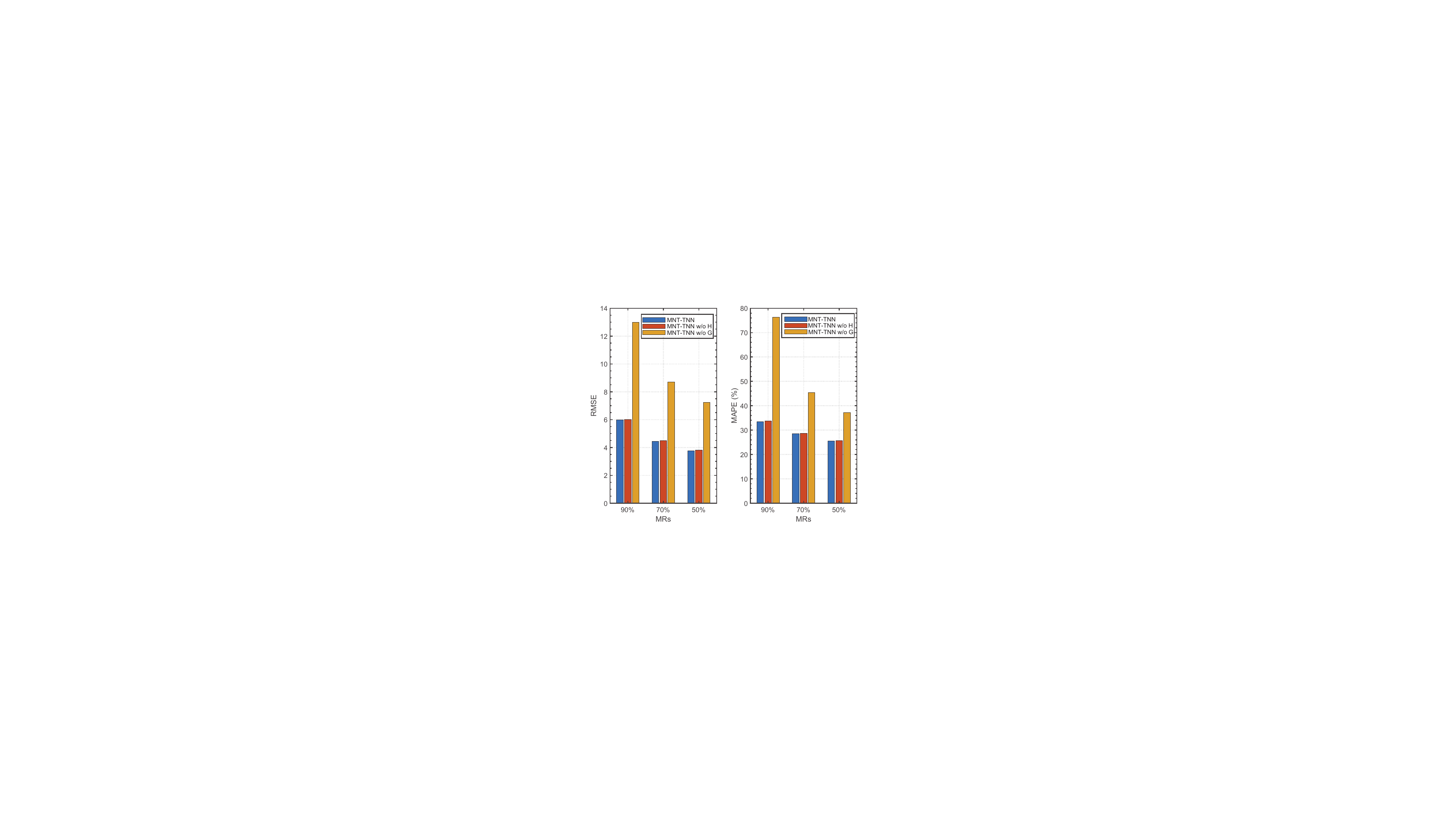}
\caption{Performances of MNT-TNN without the multiplex transform.}
\label{fig:5}
\vskip -0.1in
\end{figure}
\subsection{Ablation Study}
We evaluate the effect of the multimode transform. As shown in Section \ref{compare}, MNT-TNN achieves notable improvements in imputation accuracy over NTTNN, underscoring its effectiveness in exploring low-rankness. In this section, we remove the spatial and spatio-temporal transforms denoted by $\mathbf{G}$ and $\mathbf{H}$, respectively. {This is achieved by setting the related transform as the identity matrix and deactivating the corresponding update step in the Table. \ref{tab:2}}. Fig. \ref{fig:5} shows that both transforms strengthen the performance of MNT-TNN; however, the improvements contributed by transform $\mathbf{H}$ are marginal. {This is because for the specific problem of spatiotemporal traffic imputation with a (location × location × time) tensor structure, the spatial transform $\mathbf{G}$ combined with a temporal transform $\mathbf{T}$ already captures the necessary spatiotemporal dependencies, making an additional explicit transform $\mathbf{H}$ redundant, as analysed in Section \ref{subproblems}.} Remarkably, removing $\mathbf{G}$ causes a substantial decline in MNT-TNN's performance. This reveals not only the importance of leveraging spatial information in traffic imputation and the critical role of $\mathbf{G}$ in MNT-TNN, but also verifies our multimode low-rankness assumption for this type of tensor, i.e., each tensor mode independently contributes a specific low-rank property, and the multimode low-rankness can be captured compactly.

\begin{figure}[htbp]
% \vskip -0.2in
\centering
\includegraphics[scale=0.45]{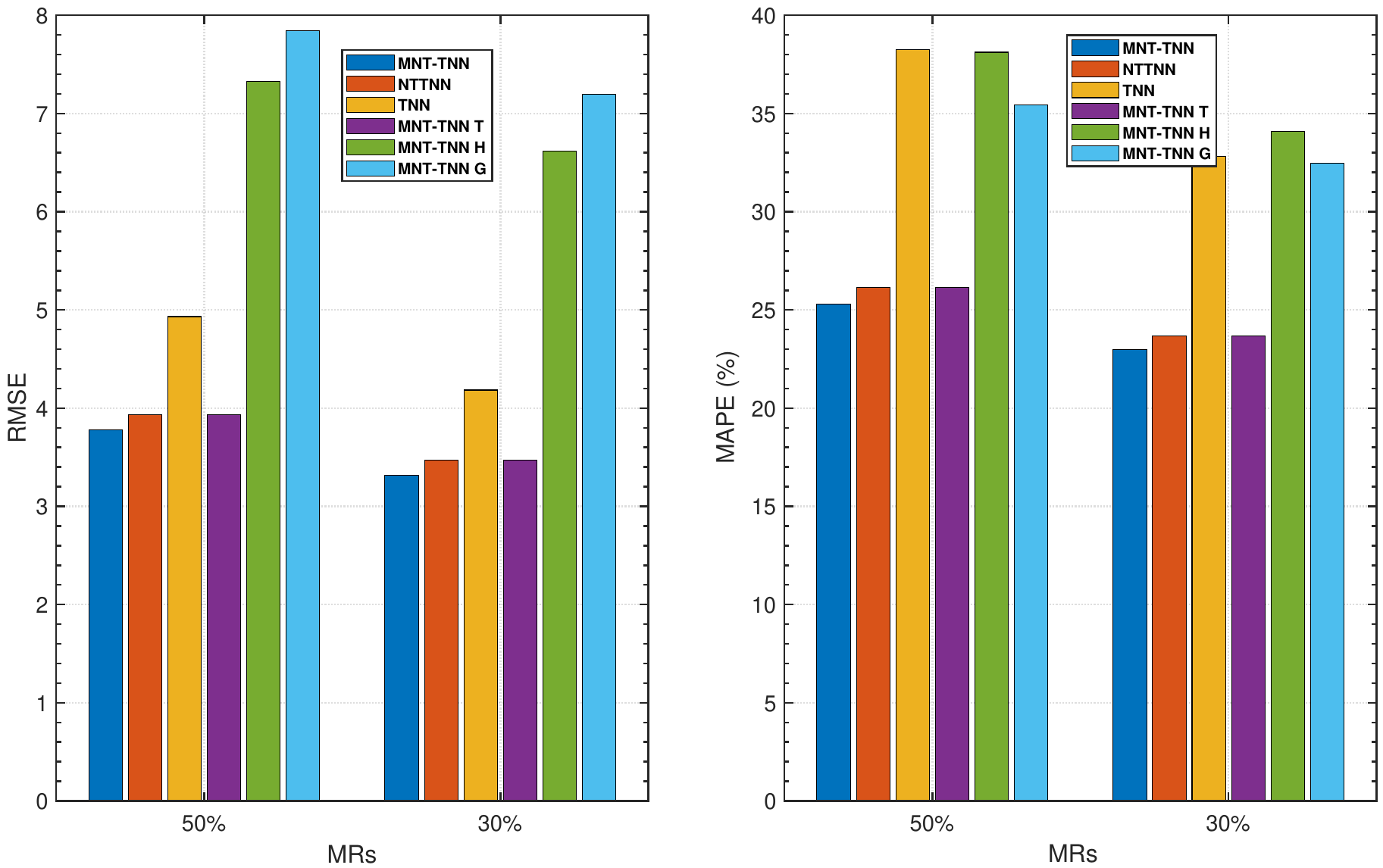}
\caption{{Comparison on TTNNs with different single generalized mode transforms.}}
\label{fig:abla_indiv}
\vskip -0.05in
\end{figure}
{To further investigate the role of the transforms, we evaluate the individual performance of each transform by removing the other two linear transforms (e.g., MNT-TNN-T denotes the version induced by only the T-transform). From the results in Fig. \ref{fig:abla_indiv}, it is noted that without the temporal transform, the imputation performance of MNT-TNN approximately degenerates to the level of the standard TNN. This result in the context provides several key insights:
(1) Under our single-mode transform strategy, MNT-TNN-T reduces to NTTNN, evidencing that MNT-TNN is indeed a generalized version of NTTNN.
(2) Although limited by the predefined transform factors, it is possible to achieve tensor recovery on any axis using the tensor nuclear norm induced by the generalized mode transform, which certifies the correctness of the proposed definitions and theorems in Sec.\ref{method}.
(3) The low-rankness and the advantage of the nonlinear function are activated primarily by the temporal mode.}
{Accordingly, we could potentially improve MNT-TNN's imputation performance by learning effective spatial and spatiotemporal transforms and adaptive weights for different modes.}

\begin{figure}[htbp]
\centering
\vskip -0.05in
\includegraphics[scale=0.75]{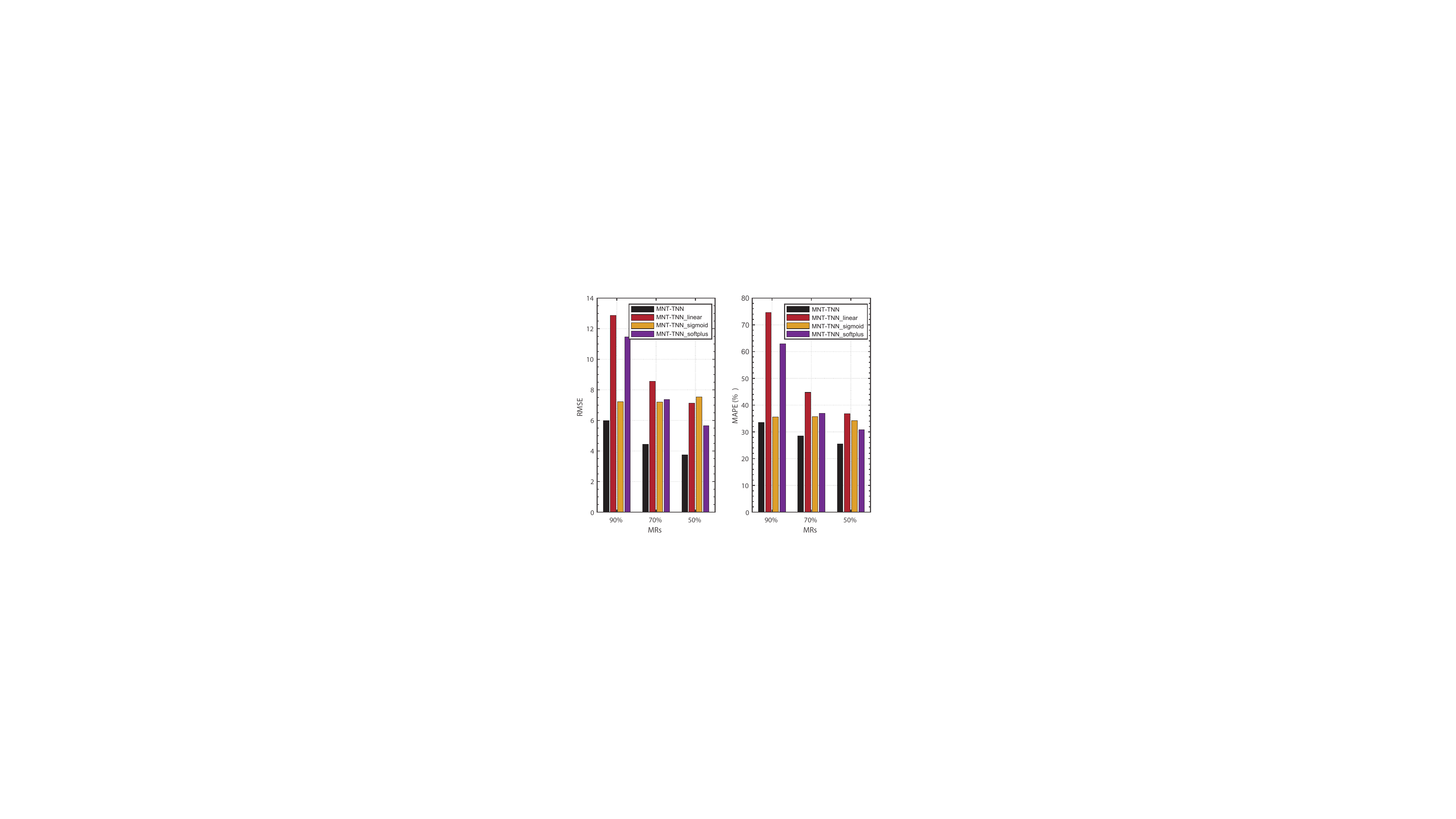}
\caption{Performances of MNT-TNN without or with different nonlinear activations.}
\label{fig:12}
\vskip -0.1in
\end{figure}
{The necessity of nonlinear activation within MNT-TNN is the subject of further examination. As demonstrated in Fig. \ref{fig:12}, the performance undergoes a substantial decline when the multimode transform is applied directly to the tensor without nonlinear activation. Furthermore, two additional nonlinear activations, sigmoid and softplus \cite{zheng2015softplus}, are employed for comparison with the Tanh function utilised in MNT-TNN. The results indicate that Tanh consistently yields optimal outcomes across a range of missing rates. This finding indicates that specific intrinsic low-rank properties inherent to the multimode of a tensor may be situated within a nonlinear-induced manifold, manifesting only under the application of suitable activations.}

\subsection{Case Study}
This part is dedicated to the sensitivity analysis of several hyperparameters concerning the performance and convergence of the proposed algorithm. All the studies in this paper are conducted on the CHSP dataset.
\begin{figure}[htbp]
% \vskip 0.2in
% \centering
\label{case_d}
\includegraphics[width=\linewidth]{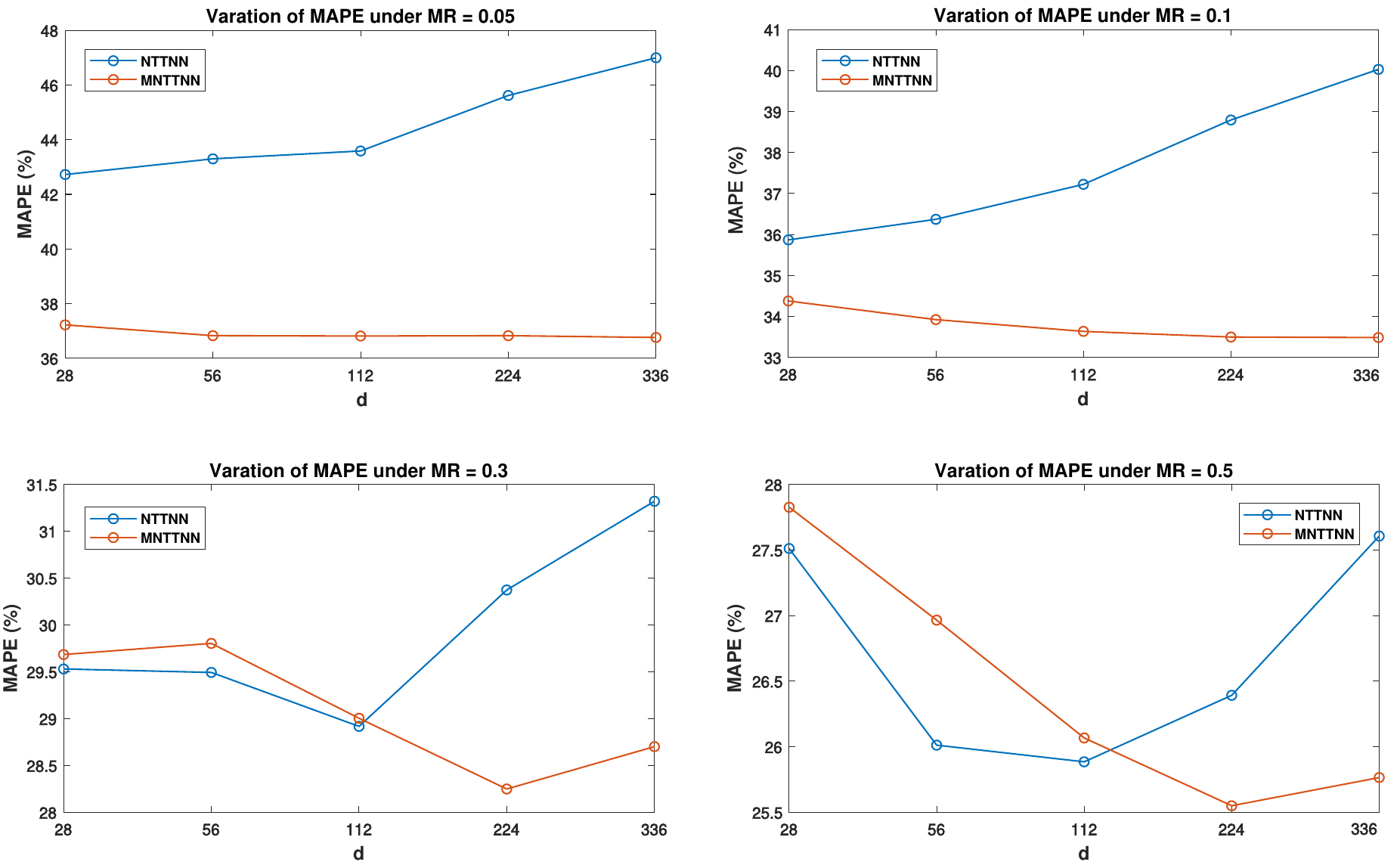}
\caption{Effects of truncated dimension d on two nonlinear TTNN methods.}
\vskip -0.1in
\end{figure}

\noindent\textbf{Effects of truncated dimension:}
In this part, we evaluate the effect of truncated dimension denoted by $d$, which is the dimension related to the truncated SVD for obtaining the semi-orthogonal transform factor $\mathbf{T} \in \mathbb{R}^{d\times m_3}$ of NTTNN and MNT-TNN. Specifically, we choose $d$ within the range of $\{28, 56, 112, 224, 336\}$, the results are shown in Fig. \ref{case_d}. Generally, we can see that MNT-TNN prefers large $d$ while NTTNN performs better with relatively small $d$. What's more, we note that MNT-TNN is more robust against the variation of $d$ when missing rates are small. 
\begin{table}[htbp]
\vskip -0.15in
\caption{Effects of the proximal coefficient on the convergence speed.}
\label{tab:5}
\centering
\begin{tabular}{c|cccccccc}
\toprule
$\rho$ & \multicolumn{2}{c}{0.001} & \multicolumn{2}{c}{0.1} & \multicolumn{2}{c}{1} & \multicolumn{2}{c}{10} \\
\cmidrule(r){2-3} \cmidrule(r){4-5} \cmidrule(r){6-7} \cmidrule(r){8-9}
MR & 10$\%$ & 50$\%$ & 10$\%$ & 50$\%$ & 10$\%$ & 50$\%$ & 10$\%$ & 50$\%$\\
\midrule
Iter & 223 & 205 & 235 & 198 & 270 & 191 & 518 & 208\\
\bottomrule
\end{tabular}
\vskip -0.1in
% \vskip 0.15in
\end{table}

\begin{figure}[htbp]
% \vskip -0.1in
% \centering
\includegraphics[width=\linewidth]{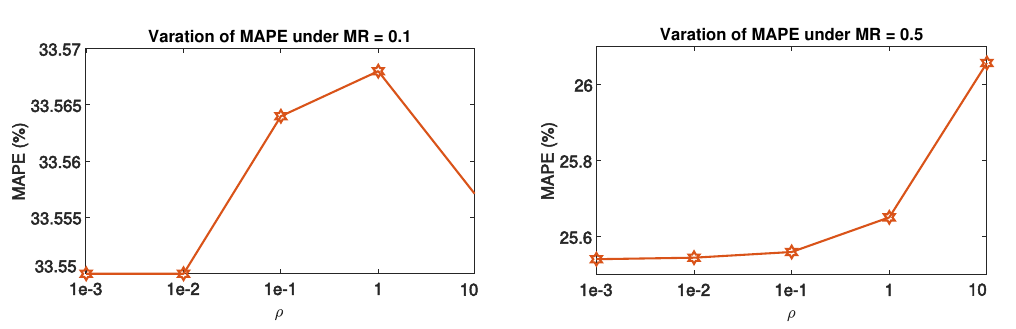}
\caption{Effects of proximal coefficient $\rho$ on the imputation performance.}
% \vskip -0.2in
\label{study_rho}
\end{figure}

\noindent\textbf{Effects of proximal coefficient:}
In this part, we study the effect of the proximal coefficient denoted by $\rho$, which determines the quality and rate of convergence of the PAM algorithm. We compare the performance of MNT-TNN across different values of $\rho$ as shown in Fig. \ref{study_rho}, which indicates that small $\rho$ values can ensure better performance. Besides, as recorded in the Table. \ref{tab:5}, large $\rho$ values may cause significant increases in iteration numbers for the convergence of the algorithm, when MR is small. For normal RMs like $50\%$, however, the number of iterations decreases marginally. To conclude, small $\rho$ values under 0.1 are recommended for our algorithm. 

\subsection{Computational Study}
\begin{figure}[htbp]
\vskip -0.2in
\centering
\includegraphics[width=.5\linewidth]{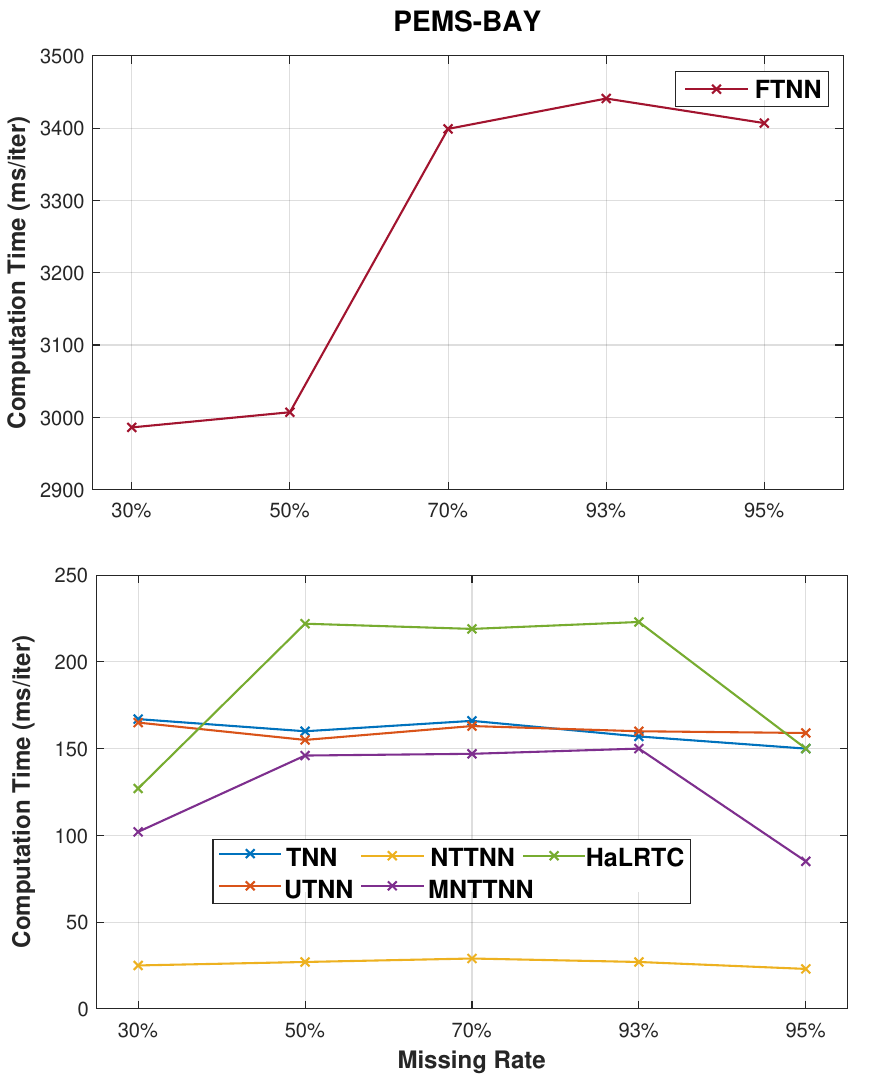}
\caption{Comparisons in terms of computation time among different optimization methods on PEMS-BAY dataset.}
\label{time:pems}
% \vskip -0.2in
\end{figure}
\begin{figure}[htbp]
% \vskip 0.2in
\centering
\includegraphics[width=.5\linewidth]{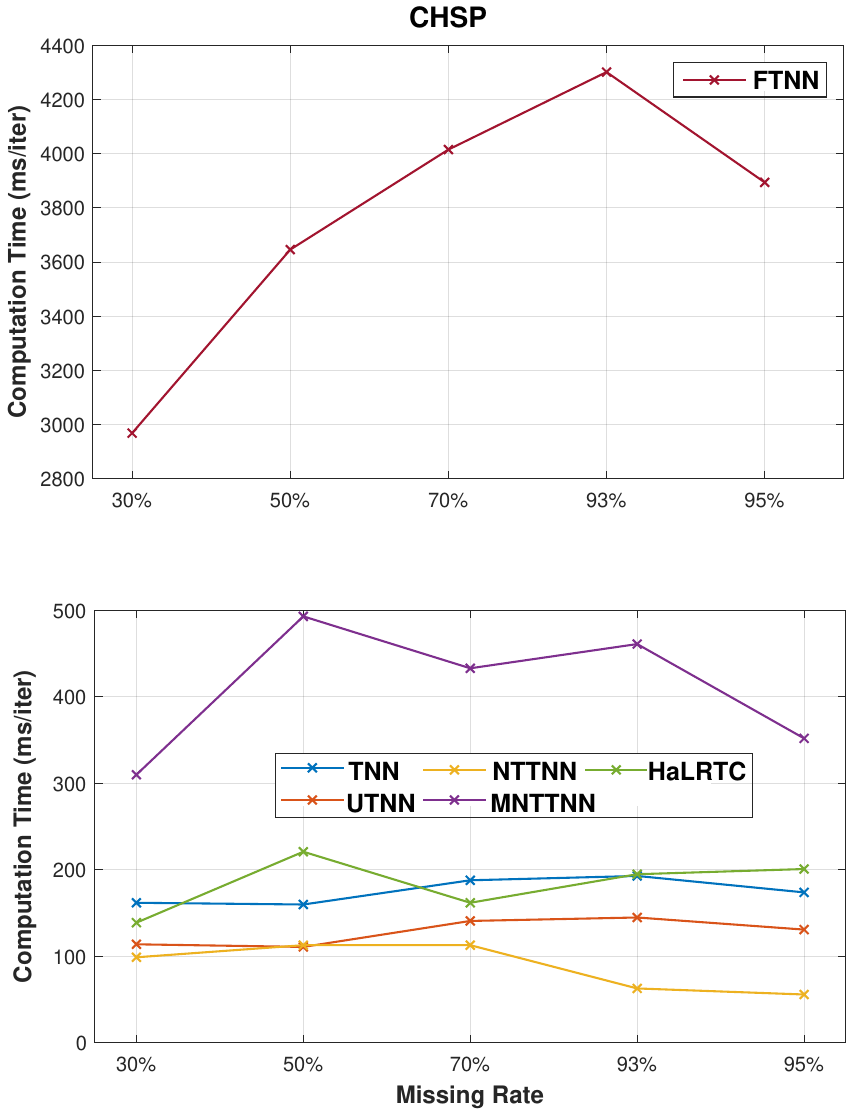}
\caption{Comparisons in terms of computation time among different optimization methods on CHSP dataset.}
\label{time:chsp}
\vskip -0.2in
\end{figure}
The imposed $\text{MNT-TNN}$ algorithm incurs additional computation costs of $O((n_1n_2)^2n_3)$ and $O(n_1^2n_3)$ flops compared with $\text{NTTNN}$ which arise from two operations, $\bar{\star}_{(1,2)} \mathbf{G}$ and $ \star_{(1,3)} \mathbf{H}$, respectively. To evaluate the time complexity of our method precisely, we record the cost time expended on one optimization iteration of different methods and present the comparison results in Fig. \ref{time:pems} and Fig. \ref{time:chsp}. Overall, NTTNN consistently outperforms other methods in terms of computation time, while FTNN needs an overwhelmingly expensive cost for each optimization iteration. The computation time of MNT-TNN varies from dataset. For the PEMS-BAY dataset, which exhibits relatively small spatial dimensions and a long temporal dimension, MNT-TNN turns out to be more efficient than all other methods except for NTTNN. However, in the case of the CHSP dataset, the efficiency of MNT-TNN suffers considerable degradation due to the quadratically increasing computation cost from an enlarged spatial dimension. 

\subsection{Further Study}
To further validate the effectiveness of MNT-TNN, we conduct supplementary experiments focusing on: (1) performance comparisons under non-random missing patterns, and (2) performance comparisons with DL-based models.
\begin{table}[h]
\vskip -0.1in
\centering
\caption{Imputation results on CHSP under the non-random missing pattern.}
\label{nonrandom}
\resizebox{\textwidth}{!}{
\begin{tabular}{c c c c c c c}
\toprule
\multirow{3}{*}{Methods} & \multicolumn{2}{c}{MR=10\%} & \multicolumn{2}{c}{MR=30\%} & \multicolumn{2}{c}{MR=50\%} \\
\cmidrule(r){2-3} \cmidrule(r){4-5} \cmidrule(r){6-7} 
 & MAPE(\%) & RMSE & MAPE(\%) & RMSE & MAPE(\%) & RMSE \\
\midrule
HaLRTC  & 38.29 & 13.66 & 56.42 & 29.35 & 78.28 & 43.73\\

LRTC-TNN & 34.25 & 10.28 & 53.33 & 26.30 & 75.68 & 40.29\\
\midrule
TNN & 47.20 & 8.41 & 56.65 & 11.25 & 86.38 & 17.11\\

UTNN & 39.82 & 7.77 & 47.85 & 10.24 & 73.74 & 14.16\\

FTNN & 40.28 & 7.81 & 46.83 & 10.83 & 74.40 & 20.30\\

NTTNN & 30.16 & 7.60 & 40.91 & 9.36 & \textbf{56.16} & 14.14\\
\midrule 

MNT-TNN & \textbf{30.16} & \textbf{7.60} & \textbf{40.91} & \textbf{9.36} & 60.20 & \textbf{13.38}\\
\bottomrule
\end{tabular}
}
\vskip -0.1in
\end{table}

First, we simulate one of the most commonly used non-random missing patterns \cite{chen2019missing}, where entire fibers along the time axis are randomly removed. The comparison results are shown in Table.\ref{nonrandom}. We observe that all optimization methods perform worse than in the random-missing case, which highlights the constraints these methods face under the random recovery theory. Nevertheless, MNT-TNN remains competitive among these methods. Additionally, we find that there is no significant advantage over NTTNN, and MNT-TNN nearly degenerates into NTTNN when the missing rate is low. This suggests that the spatial transform does not contribute effectively in this scenario. {One likely reason is that, under such a non-random missing pattern, the effective spatial dependencies become extremely sparse due to the missing fibers. This would lead to an incompatibility between the incomplete spatial graph and the applied spatial transform since the graph Laplacian is constructed based on the full set of nodes.}

Second, we conduct further experiments comparing two recent representative deep-learning methods for spatiotemporal imputation on our datasets. One is CSDI \cite{tashiro2021csdi}, a powerful imputation method based on the diffusion model \cite{ho2020denoising}. The other is Imputeformer \cite{nie2024imputeformer}, a low-rankness-based imputation network built on transformers \cite{vaswani2017attention}, which has recently achieved state-of-the-art results on a wide range of spatiotemporal datasets. To meet the training and validation requirements for these models, we reorganize the spatiotemporal tensor into the form $\text{batch}\times \text{node}\times \text{time}$, based on the size of the dataset. For example, the CHSP tensor, originally of size $30 \times30 \times528$, is reshaped into batches of size $16 \times900 \times33$. We use the first twelve batches for training and the last four batches for testing. Final results are obtained via k-fold cross-validation.
\begin{table}[htbp]
\label{dl-chsp}
\vskip -0.1in
\centering
\caption{Imputation results compared with the DL-based methods on CHSP.}
\vskip 0.1in
\begin{tabular}{c c c c}
\toprule
method & missing rate & RMSE & MAE \\
\midrule
\multirow{3}{*}{CSDI} & 50\% & 11.55 & 4.78 \\
& 70\% & 12.04 & 5.75\\
& 90\% & 14.84 & 7.88\\
\cmidrule{2-4}
\multirow{3}{*}{ImputeFormer}& 50\% & 25.51 & 23.45\\
& 70\% & 30.12 & 29.10\\
& 90\% & 34.32 & 34.30\\
\midrule
\multirow{3}{*}{NTTNN} & 50\% & 3.94 & 2.67\\
& 70\% & 4.70 & 3.16\\
& 90\% & 6.39 & 4.11\\
\cmidrule{2-4}
\multirow{3}{*}{MNTTNN} & 50\% & \textbf{3.77} & \textbf{2.57}\\
& 70\% & \textbf{4.47} & \textbf{3.02}\\
& 90\% & \textbf{6.10} & \textbf{3.98}\\
\bottomrule
\end{tabular}
\end{table}

\begin{table}[htbp]
\vskip -0.1in
\centering
\caption{Imputation results compared with the DL-based methods on PEMS.}
\vskip 0.1in
\begin{tabular}{c c c c}
\toprule
method & missing rate & RMSE & MAE \\
\midrule
\multirow{3}{*}{CSDI} & 50\% & 27.85 & 21.85 \\
& 70\% & 30.59 & 24.42\\
& 90\% & 33.37 & 26.76\\
\cmidrule{2-4}
\multirow{3}{*}{ImputeFormer}& 50\% & 28.28 & 27.05 \\
& 70\% & 32.84 & 32.71\\
& 90\% & 35.98 & 34.78\\
\midrule
\multirow{3}{*}{NTTNN} & 50\% &  27.05 & 17.27 \\
& 70\% & 27.82 & 17.81\\
& 90\% & 33.47 & 22.20\\
\cmidrule{2-4}
\multirow{3}{*}{MNTTNN} & 50\% & \textbf{26.33} & \textbf{16.66} \\
& 70\% & \textbf{27.75} & \textbf{17.60} \\
& 90\% & \textbf{33.27} & \textbf{21.50} \\
\bottomrule
\end{tabular}
\end{table}

As reported in the Tables.\ref{dl-chsp}, DL models consistently demonstrate poorer performance and instability in imputation tasks on the datasets under consideration. These results are not surprising, as discussed in Section. \ref{intro}, the demand for large volumes of training data and the need for extensive tuning of model hyperparameters and training settings make it difficult and time-consuming to fully explore the capabilities of these methods. This limits their practical effectiveness, further highlighting the robustness of the proposed method in many real-world scenarios.

\section{Limitation Analysis}
\label{limitation}
Based on our experimental analysis, we identify several areas for potential improvement in the MNT-TNN method.

First, the improvements achieved by $\text{MNT-TNN}$ appear restricted due to a few factors: (1) The representation capacity is limited as the transform factors are pre-constructed and fixed. (2) As a nonlinear optimization method, MNT-TNN is sometimes sensitive to initializations and hyperparameters, which is an issue partly mitigated by \text{ATTNNs}. 

Second, owing to the variations in dataset size and tensor structure, the improvements achieved by MNT-TNN on certain datasets, such as PEMS, are marginal. This, on the one hand, indicates the benefits of leveraging diverse modalities; On the other hand, it highlights certain limitations. Specifically, while the PEMS tensor displays robust temporal correlations, its sparse spatial patterns contribute limited spatial dynamics, resulting in few performance gains despite the additional computational burden from the multi-mode transform. Thus, to balance efficiency and effectiveness, we suggest that our method be applied selectively to tensors with substantial spatial dependencies and moderate spatial dimensions (n1 and n2 are relatively small).

Finally, despite the robustness provided by the tensor recovery theorems, the recovery performances are limited in the case of non-random missing, which also limits the application of this method to other practical imputation scenarios. How to combine the advantage of the multimode transform-based optimization method with modern techniques such as deep learning models to improve the imputation performance and broaden the range of applications is an attractive and promising direction for future research.

\section{Conclusion and Future Work}
\label{conclusion}
In this paper, {we propose a general Multimode Nonlinear Transformation-based Tensor Nuclear Norm (MNT-TNN) and apply it for the specific problem} of randomly missing values in the imputation of spatiotemporal traffic data. It is based on a general Multimode Nonlinear Transform (MNT) and TTNN framework. We address the nonconvex optimization problem by applying the Proximal Alternating Minimization (PAM) algorithm with theoretical convergence guarantees. Furthermore, we propose an Augmented TTNN Families (ATTNNs) framework that uses various TTNN techniques to improve imputation performance under high missing rates. We conduct extensive experiments to evaluate the effectiveness of the proposed MNT-TNN method, which consistently outperforms other compared methods. In the future, we will focus on developing more efficient implementations for MNT-TNN optimization, extending its application to other tensor data types such as images and videos, {and exploring learnable transform factors for exploiting multimode low-rankness of tensors in both random and non-random cases}. 

\bibliography{references}

\begin{thebibliography}{10}
\expandafter\ifx\csname url\endcsname\relax
  \def\url#1{\texttt{#1}}\fi
\expandafter\ifx\csname urlprefix\endcsname\relax\def\urlprefix{URL }\fi
\expandafter\ifx\csname href\endcsname\relax
  \def\href#1#2{#2} \def\path#1{#1}\fi

\bibitem{zheng2020gman}
C.~Zheng, X.~Fan, C.~Wang, J.~Qi, Gman: A graph multi-attention network for traffic prediction, in: Proceedings of the AAAI conference on artificial intelligence, Vol.~34, 2020, pp. 1234--1241.

\bibitem{traffdetect}
L.~Deng, D.~Lian, Z.~Huang, E.~Chen, Graph convolutional adversarial networks for spatiotemporal anomaly detection, IEEE Transactions on Neural Networks and Learning Systems 33~(6) (2022) 2416--2428.
\newblock \href {https://doi.org/10.1109/TNNLS.2021.3136171} {\path{doi:10.1109/TNNLS.2021.3136171}}.

\bibitem{wang2023human}
D.~Wang, L.~Wu, D.~Zhang, J.~Zhou, L.~Sun, Y.~Fu, Human-instructed deep hierarchical generative learning for automated urban planning, in: Proceedings of the AAAI Conference on Artificial Intelligence, Vol.~37, 2023, pp. 4660--4667.

\bibitem{wang2018adaptivespatialtemporal}
Y.~Wang, Y.~Zhang, X.~Piao, H.~Liu, K.~Zhang, Traffic data reconstruction via adaptive spatial-temporal correlations, IEEE Transactions on Intelligent Transportation Systems 20~(4) (2018) 1531--1543.

\bibitem{miao2021generative}
X.~Miao, Y.~Wu, J.~Wang, Y.~Gao, X.~Mao, J.~Yin, Generative semi-supervised learning for multivariate time series imputation, in: Proceedings of the AAAI conference on artificial intelligence, Vol.~35, 2021, pp. 8983--8991.

\bibitem{tashiro2021csdi}
Y.~Tashiro, J.~Song, Y.~Song, S.~Ermon, Csdi: Conditional score-based diffusion models for probabilistic time series imputation, Advances in Neural Information Processing Systems 34 (2021) 24804--24816.

\bibitem{nie2024imputeformer}
T.~Nie, G.~Qin, W.~Ma, Y.~Mei, J.~Sun, Imputeformer: Low rankness-induced transformers for generalizable spatiotemporal imputation, in: Proceedings of the 30th ACM SIGKDD Conference on Knowledge Discovery and Data Mining, 2024, pp. 2260--2271.

\bibitem{yu2016temporal}
H.-F. Yu, N.~Rao, I.~S. Dhillon, Temporal regularized matrix factorization for high-dimensional time series prediction, Advances in neural information processing systems 29 (2016).

\bibitem{zhu2012compressive}
Y.~Zhu, Z.~Li, H.~Zhu, M.~Li, Q.~Zhang, A compressive sensing approach to urban traffic estimation with probe vehicles, IEEE Transactions on Mobile Computing 12~(11) (2012) 2289--2302.

\bibitem{qu2009ppca}
L.~Qu, L.~Li, Y.~Zhang, J.~Hu, Ppca-based missing data imputation for traffic flow volume: A systematical approach, IEEE Transactions on intelligent transportation systems 10~(3) (2009) 512--522.

\bibitem{li2013efficient}
L.~Li, Y.~Li, Z.~Li, Efficient missing data imputing for traffic flow by considering temporal and spatial dependence, Transportation research part C: emerging technologies 34 (2013) 108--120.

\bibitem{yu2020urban}
J.~Yu, M.~E. Stettler, P.~Angeloudis, S.~Hu, X.~M. Chen, Urban network-wide traffic speed estimation with massive ride-sourcing gps traces, Transportation Research Part C: Emerging Technologies 112 (2020) 136--152.

\bibitem{chen2018spatialsvd}
X.~Chen, Z.~He, J.~Wang, Spatial-temporal traffic speed patterns discovery and incomplete data recovery via svd-combined tensor decomposition, Transportation research part C: emerging technologies 86 (2018) 59--77.

\bibitem{chen2021latc}
X.~Chen, M.~Lei, N.~Saunier, L.~Sun, Low-rank autoregressive tensor completion for spatiotemporal traffic data imputation, IEEE Transactions on Intelligent Transportation Systems 23~(8) (2021) 12301--12310.

\bibitem{chen2021lstc}
X.~Chen, Y.~Chen, N.~Saunier, L.~Sun, Scalable low-rank tensor learning for spatiotemporal traffic data imputation, Transportation research part C: emerging technologies 129 (2021) 103226.

\bibitem{hu2020robust}
Y.~Hu, D.~B. Work, Robust tensor recovery with fiber outliers for traffic events, ACM Transactions on Knowledge Discovery from Data (TKDD) 15~(1) (2020) 1--27.

\bibitem{ni2007determiningITS}
D.~Ni, Determining traffic-flow characteristics by definition for application in its, IEEE Transactions on Intelligent Transportation Systems 8~(2) (2007) 181--187.

\bibitem{LuFCLLY20}
C.~Lu, J.~Feng, Y.~Chen, W.~Liu, Z.~Lin, S.~Yan, \href{https://doi.org/10.1109/TPAMI.2019.2891760}{Tensor robust principal component analysis with a new tensor nuclear norm}, {IEEE} Trans. Pattern Anal. Mach. Intell. 42~(4) (2020) 925--938.
\newblock \href {https://doi.org/10.1109/TPAMI.2019.2891760} {\path{doi:10.1109/TPAMI.2019.2891760}}.
\newline\urlprefix\url{https://doi.org/10.1109/TPAMI.2019.2891760}

\bibitem{ZhangA17}
Z.~Zhang, S.~Aeron, \href{https://doi.org/10.1109/TSP.2016.2639466}{Exact tensor completion using t-svd}, {IEEE} Trans. Signal Process. 65~(6) (2017) 1511--1526.
\newblock \href {https://doi.org/10.1109/TSP.2016.2639466} {\path{doi:10.1109/TSP.2016.2639466}}.
\newline\urlprefix\url{https://doi.org/10.1109/TSP.2016.2639466}

\bibitem{asif2013cptraffic}
M.~T. Asif, N.~Mitrovic, L.~Garg, J.~Dauwels, P.~Jaillet, Low-dimensional models for missing data imputation in road networks, in: 2013 IEEE International Conference on Acoustics, Speech and Signal Processing, IEEE, 2013, pp. 3527--3531.

\bibitem{yokota2016smoothcp}
T.~Yokota, Q.~Zhao, A.~Cichocki, Smooth parafac decomposition for tensor completion, IEEE Transactions on Signal Processing 64~(20) (2016) 5423--5436.

\bibitem{erichson2020randomizedcp}
N.~B. Erichson, K.~Manohar, S.~L. Brunton, J.~N. Kutz, Randomized cp tensor decomposition, Machine Learning: Science and Technology 1~(2) (2020) 025012.

\bibitem{tan2013tucker}
H.~Tan, G.~Feng, J.~Feng, W.~Wang, Y.-J. Zhang, F.~Li, A tensor-based method for missing traffic data completion, Transportation Research Part C: Emerging Technologies 28 (2013) 15--27.

\bibitem{goulart2017traffictuc}
J.~d.~M. Goulart, A.~Kibangou, G.~Favier, Traffic data imputation via tensor completion based on soft thresholding of tucker core, Transportation Research Part C: Emerging Technologies 85 (2017) 348--362.

\bibitem{gandy2011lownranktucker}
S.~Gandy, B.~Recht, I.~Yamada, Tensor completion and low-n-rank tensor recovery via convex optimization, Inverse problems 27~(2) (2011) 025010.

\bibitem{hillar2013nphard}
C.~J. Hillar, L.-H. Lim, Most tensor problems are np-hard, Journal of the ACM (JACM) 60~(6) (2013) 1--39.

\bibitem{nie2023tensorsvd}
T.~Nie, G.~Qin, Y.~Wang, J.~Sun, Correlating sparse sensing for large-scale traffic speed estimation: A laplacian-enhanced low-rank tensor kriging approach, Transportation research part C: emerging technologies 152 (2023) 104190.

\bibitem{deng2021graph}
L.~Deng, X.-Y. Liu, H.~Zheng, X.~Feng, Y.~Chen, Graph spectral regularized tensor completion for traffic data imputation, IEEE Transactions on Intelligent Transportation Systems 23~(8) (2021) 10996--11010.

\bibitem{oseledets2011tensortrain}
I.~V. Oseledets, Tensor-train decomposition, SIAM Journal on Scientific Computing 33~(5) (2011) 2295--2317.

\bibitem{zheng2021tensornet}
Y.-B. Zheng, T.-Z. Huang, X.-L. Zhao, Q.~Zhao, T.-X. Jiang, Fully-connected tensor network decomposition and its application to higher-order tensor completion, in: Proceedings of the AAAI conference on artificial intelligence, Vol.~35, 2021, pp. 11071--11078.

\bibitem{liu2012lrtc}
J.~Liu, P.~Musialski, P.~Wonka, J.~Ye, Tensor completion for estimating missing values in visual data, IEEE transactions on pattern analysis and machine intelligence 35~(1) (2012) 208--220.

\bibitem{tan2014trafficlrtc}
H.~Tan, W.~Yuankai, J.~Feng, W.~Wang, B.~Ran, Traffic missing data completion with spatial-temporal correlations, Department of Civil and Environmental Engineering, University of Wisconsin-Madison (2014).

\bibitem{chen2020nonconvex}
X.~Chen, J.~Yang, L.~Sun, A nonconvex low-rank tensor completion model for spatiotemporal traffic data imputation, Transportation Research Part C: Emerging Technologies 117 (2020) 102673.

\bibitem{nie2022schattenp}
T.~Nie, G.~Qin, J.~Sun, Truncated tensor schatten p-norm based approach for spatiotemporal traffic data imputation with complicated missing patterns, Transportation research part C: emerging technologies 141 (2022) 103737.

\bibitem{kernfeld2015tensorprod}
E.~Kernfeld, M.~Kilmer, S.~Aeron, Tensor--tensor products with invertible linear transforms, Linear Algebra and its Applications 485 (2015) 545--570.

\bibitem{kilmer2021tenalgebra}
M.~E. Kilmer, L.~Horesh, H.~Avron, E.~Newman, Tensor-tensor algebra for optimal representation and compression of multiway data, Proceedings of the National Academy of Sciences 118~(28) (2021) e2015851118.

\bibitem{DBLP:journals/cacm/CandesR12}
E.~J. Cand{\`{e}}s, B.~Recht, Exact matrix completion via convex optimization, Commun. {ACM} 55~(6) (2012) 111--119.

\bibitem{SongNZ20}
G.~Song, M.~K. Ng, X.~Zhang, \href{https://doi.org/10.1002/nla.2299}{Robust tensor completion using transformed tensor singular value decomposition}, Numer. Linear Algebra Appl. 27~(3) (2020).
\newblock \href {https://doi.org/10.1002/NLA.2299} {\path{doi:10.1002/NLA.2299}}.
\newline\urlprefix\url{https://doi.org/10.1002/nla.2299}

\bibitem{JiangNZH20}
T.~Jiang, M.~K. Ng, X.~Zhao, T.~Huang, \href{https://doi.org/10.1109/TIP.2020.3000349}{Framelet representation of tensor nuclear norm for third-order tensor completion}, {IEEE} Trans. Image Process. 29 (2020) 7233--7244.
\newblock \href {https://doi.org/10.1109/TIP.2020.3000349} {\path{doi:10.1109/TIP.2020.3000349}}.
\newline\urlprefix\url{https://doi.org/10.1109/TIP.2020.3000349}

\bibitem{wang2022conotcoupled}
J.-L. Wang, T.-Z. Huang, X.-L. Zhao, Y.-S. Luo, T.-X. Jiang, Conot: Coupled nonlinear transform-based low-rank tensor representation for multidimensional image completion, IEEE Transactions on Neural Networks and Learning Systems (2022).

\bibitem{LiZJZH22}
B.~Li, X.~Zhao, T.~Ji, X.~Zhang, T.~Huang, \href{https://doi.org/10.1007/s10915-022-01937-1}{Nonlinear transform induced tensor nuclear norm for tensor completion}, J. Sci. Comput. 92~(3) (2022) 83.
\newblock \href {https://doi.org/10.1007/S10915-022-01937-1} {\path{doi:10.1007/S10915-022-01937-1}}.
\newline\urlprefix\url{https://doi.org/10.1007/s10915-022-01937-1}

\bibitem{ZhangEAHK14}
Z.~Zhang, G.~Ely, S.~Aeron, N.~Hao, M.~E. Kilmer, \href{https://doi.org/10.1109/CVPR.2014.485}{Novel methods for multilinear data completion and de-noising based on tensor-svd}, in: 2014 {IEEE} Conference on Computer Vision and Pattern Recognition, {CVPR} 2014, Columbus, OH, USA, June 23-28, 2014, {IEEE} Computer Society, 2014, pp. 3842--3849.
\newblock \href {https://doi.org/10.1109/CVPR.2014.485} {\path{doi:10.1109/CVPR.2014.485}}.
\newline\urlprefix\url{https://doi.org/10.1109/CVPR.2014.485}

\bibitem{krishnan2009fast}
D.~Krishnan, R.~Fergus, Fast image deconvolution using hyper-laplacian priors, Advances in neural information processing systems 22 (2009).

\bibitem{cai2010singular}
J.-F. Cai, E.~J. Cand{\`e}s, Z.~Shen, A singular value thresholding algorithm for matrix completion, SIAM Journal on optimization 20~(4) (2010) 1956--1982.

\bibitem{schonemann1966generalized}
P.~H. Sch{\"o}nemann, A generalized solution of the orthogonal procrustes problem, Psychometrika 31~(1) (1966) 1--10.

\bibitem{chen2013tensorfactor}
Y.-L. Chen, C.-T. Hsu, H.-Y.~M. Liao, Simultaneous tensor decomposition and completion using factor priors, IEEE Transactions on Pattern Analysis and Machine Intelligence 36~(3) (2014) 577--591.
\newblock \href {https://doi.org/10.1109/TPAMI.2013.164} {\path{doi:10.1109/TPAMI.2013.164}}.

\bibitem{pan2021spatiotemporal}
C.~Pan, S.~Chen, A.~Ortega, Spatio-temporal graph scattering transform, arXiv preprint arXiv:2012.03363 (2021).

\bibitem{liu2022scinet}
M.~Liu, A.~Zeng, M.~Chen, Z.~Xu, Q.~Lai, L.~Ma, Q.~Xu, Scinet: Time series modeling and forecasting with sample convolution and interaction, Advances in Neural Information Processing Systems 35 (2022) 5816--5828.

\bibitem{yair2018multi}
N.~Yair, T.~Michaeli, Multi-scale weighted nuclear norm image restoration, in: Proceedings of the IEEE conference on computer vision and pattern recognition, 2018, pp. 3165--3174.

\bibitem{shuman2013emerging}
D.~I. Shuman, S.~K. Narang, P.~Frossard, A.~Ortega, P.~Vandergheynst, The emerging field of signal processing on graphs: Extending high-dimensional data analysis to networks and other irregular domains, IEEE signal processing magazine 30~(3) (2013) 83--98.

\bibitem{lim2020hodge}
L.-H. Lim, Hodge laplacians on graphs, Siam Review 62~(3) (2020) 685--715.

\bibitem{zheng2015softplus}
H.~Zheng, Z.~Yang, W.~Liu, J.~Liang, Y.~Li, Improving deep neural networks using softplus units, in: 2015 International joint conference on neural networks (IJCNN), IEEE, 2015, pp. 1--4.

\bibitem{chen2019missing}
X.~Chen, Z.~He, Y.~Chen, Y.~Lu, J.~Wang, Missing traffic data imputation and pattern discovery with a bayesian augmented tensor factorization model, Transportation Research Part C: Emerging Technologies 104 (2019) 66--77.

\bibitem{ho2020denoising}
J.~Ho, A.~Jain, P.~Abbeel, Denoising diffusion probabilistic models, Advances in neural information processing systems 33 (2020) 6840--6851.

\bibitem{vaswani2017attention}
A.~Vaswani, N.~Shazeer, N.~Parmar, J.~Uszkoreit, L.~Jones, A.~N. Gomez, {\L}.~Kaiser, I.~Polosukhin, Attention is all you need, Advances in neural information processing systems 30 (2017).

\bibitem{attouch2010proximal}
H.~Attouch, J.~Bolte, P.~Redont, A.~Soubeyran, Proximal alternating minimization and projection methods for nonconvex problems: An approach based on the kurdyka-{\l}ojasiewicz inequality, Mathematics of operations research 35~(2) (2010) 438--457.

\bibitem{bolte2014proximal}
J.~Bolte, S.~Sabach, M.~Teboulle, Proximal alternating linearized minimization for nonconvex and nonsmooth problems, Mathematical Programming 146~(1-2) (2014) 459--494.

\bibitem{attouch2013convergence}
H.~Attouch, J.~Bolte, B.~F. Svaiter, Convergence of descent methods for semi-algebraic and tame problems: proximal algorithms, forward--backward splitting, and regularized gauss--seidel methods, Mathematical Programming 137~(1-2) (2013) 91--129.

\end{thebibliography}
% \section{Declaration of generative AI and AI-assisted technologies in the writing process}

% Statement: During the preparation of this work the author(s) used ChatGPT in order to refine the sentences and expressions. After using this tool/service, the author(s) reviewed and edited the content as needed and take(s) full responsibility for the content of the published article.
% \section*{Acknowledgments}
% This research is jointly supported by the Presidential Foundation of the Chinese Academy of Sciences, China (Grant No.YZJJ202301-CX) and the project supported by the Anhui Province Natural Science foundation, China (Grant No.2308085US01).

\appendix
\section{Further Explanations on Tensor Operations}
\label{app:tq}
\begin{definition}[Generalized Mode Unfolding (GMU)]
Given a 3D real tensor $\mathcal{X}\in \mathbb{R}^{n_1\times n_2\times n_3}$, the GMU is given as
\begin{equation}
% \label{eq:2}
    \mathbf{Unfold}(\mathcal{X},S):=\mathbf{X}_{[S]} \in \mathbb{R}^{{\times}_{i\in S} n_i{\times} (\prod_{j\in N,j\notin S} n_j)}
\end{equation}
where $S$ is an ordered subset of the indices's set $N = \{1,2,3\}$. The symbol ${\times}$ denotes the Cartesian product, which should be distinguished from the product of numbers represented by $\prod$.
The inverse operation of unfolding is represented by  
\begin{equation}
\mathbf{Fold}_{S}(X_{[S]}) := \mathcal{X}
\end{equation}
In addition, we can define a variant of this unfolding,
\begin{equation}
% \label{eq:4}
\mathbf{\overline{X}}_{[S]}\in \mathbb{R}^{\prod_{i\in S}n_i{\times} \prod_{j\in N,j\notin S}n_j}
\end{equation}
\end{definition}
\begin{remark}
    Denote $C(S)$ to be the cardinality of $S$. Note that when $C(S)$ equals 1, the new unfolding achieved by Eq. (\ref{eq:2}) reduces to the normal mode-$k$ unfolding; and when $C(S)=2$, it can additionally represent the operation of dimensionality rearrangement; In the meanwhile, Eq. (\ref{eq:4}) basically represents the vectorization of a tensor as $C(S) = 3$. As we will demonstrate, this variant plays a crucial role in the spatial transform within our proposed MNT-TNN method.
\end{remark}
\begin{definition}[2D Mode Product]
Suppose a real matrix $\mathbf{M}\in \mathbb{R}^{m \times n_k}$, recall that the mode-$k$ product of tensor $\mathcal{X}$ with respect to $\mathbf{M}$ is defined as  $\mathcal{X} \times_k \mathbf{M}=\mathbf{Fold}_{(k)}(\mathbf{M\mathbf{X}}_{(k)}),~k\in N$. In an analogous way, we define a 2D mode product for an arbitrary tensor with respect to any linear operator, involving two similar algebraic operations as follows:
    
By setting $C(S)=2$ in Eq. (\ref{eq:2}), we define the mode-($k,p$) product of tensor $\mathcal{X}$ with respect to the matrix $\mathbf{M}\in \mathbb{R}^{m\times n_k}$ as the form of the so-called face-wise product as 
\begin{gather}
% \label{eq:2dtrans}
(\mathcal{X}{\star}_{(k,p)} \mathbf{M})^{(i)}:=\mathbf{M}\mathbf{X}_{[(k,p)]}^{(i)},\\
\notag k,p~\in N,i=1,2,\dots,\prod \limits_{j\in N,j\neq k,p}n_j
\end{gather}
\begin{remark}
    One can notice that when $C(S)=1$, it degenerates to the normal mode-$k$ product after removing the superscript $(i)$ and changing $\star$ into $\times$. Notably, for 3D tensors, the effect of applying this product is equivalent to the mode-$k$ product.
\end{remark} 

Let the factor matrix $\mathbf{M}$ belong to $\mathbb{R}^{m\times n_kn_p}$. The variant of this product is defined by first applying an additional vectorization to each slice to be multiplied,
\begin{equation}
\mathcal{X}~{\bar\star}_{(k,p)}\mathbf{M}:=\mathbf{Fold}_{(k,p)}(\mathbf{M}\overline{\mathbf{X}}_{[(k,p)]})
\end{equation}
\end{definition}
\begin{remark}
These products are based entirely on variations in shape and linear algebra within the vector space, ensuring that they are well-defined. Analogous to the mode-$k$ product, which acts as a 1D transformation applied to each fiber-tube along the $k$-th mode, the 2D mode product in Eq. (\ref{eq:2dtrans}) is precisely a transformation applied to each frontal slice along a specified mode.
\end{remark}
\begin{definition}[Multimode Nonlinear Transform (MNT)]
 For any tensor $\mathcal{X}\in \mathbb{R}^{n_1\times n_2\times n_3}$, the MNT is defined as 
\begin{equation}
\mathcal{C} = \psi(\mathcal{X}~\bar\star_{p\in P} \mathbf{U}_p \star_{q\in Q} \mathbf{U}_q),~~P,Q\subseteq 2^N
\end{equation}
where $\psi(\cdot)$ is a specified element-wise function, $2^N$ is the power set of $N$, and $P, Q$ are ordered subsets. Unlike the Tucker decomposition described in Eq. (\ref{eq:tucker}), here $\mathbf{U}_p$ and $\mathbf{U}_q$ can be relaxed to semi-orthogonal matrices. We refer to the resulting tensor $\mathcal{C}$ as the transformed kernel/core of the original tensor. 
\end{definition}
\begin{remark}
Although $N$ is unique and fixed, the linear transform applied to any mode can be composed as many times as possible by independently performing matrix multiplications. As nonlinear activation has been shown to be effective in enhancing the low-rankness exploration \cite{LiZJZH22}, we incorporate this trick into our method, allowing for any number of activations between two linear transforms. Importantly, because all linear operators are semi-orthogonal, the concept of transformed kernel is valid in both directions. Concretely, setting $\psi(x)$ to be an identity function and defining $P$,$Q$ to be the empty set and $N$ respectively reduces this to the Tucker decomposition from $\mathcal{C}$ to $\mathcal{X}$, and vice versa.
\end{remark}
\section{Convergence Analysis of MNT-TNN}
\label{app:conv}
Under the framework of PAM, we establish the convergence of the solution sequence obtained by the process Eq. (\ref{eq:process}) for solving the problem Eq. (\ref{eq:prob})

First of all, we introduce the following definitions:
\begin{definition}[Kurdyka-Łojasiewicz property \cite{attouch2010proximal}]
The function $f$ is said to own the the Kurdyka-Łojasiewicz property at $\bar x \in dom(\partial f)$ if there exist $\eta \in \left(0, +\infty \right]$, a neighborhood $U$ of $\bar x$ and a continuous concave function $\varphi(x):\left(0,\eta \right]\to \mathbb{R}_{+}$, such that 

(i) $\varphi(0) = 0$,

(ii) $\varphi$ is $C^1$ on $(0, \eta)$,

(iii) for all $x\in(0,\eta)$, $\varphi^\prime(x)\ge 0$,

(iv) for all $x$ in $U\cap\{x\in \mathbb{R}^n:f(\bar x)\le f(x) < f(\bar x)+\eta \}$, the following Kurdyka-Łojasiewicz inequality holds:
\begin{equation*}
\varphi^\prime(f(x) -f(\bar x))~\text{dist}(0, \partial f(x))\ge 1
\end{equation*}
where $\text{dist}(x, K)$ is the distance from any point $x\in \mathbb{R}^n$ to the subset $K$ of $\mathbb{R}^n$. The proper lower semi-continuous functions are called K-Ł functions if they satisfy the K-Ł property at every point of $dom(\partial f)$.
\end{definition}

\begin{definition}[semialgebraic set and semialgebraic function \cite{bolte2014proximal}]
The subset $S\in \mathbb{R}$ is a semialgebraic set if there exist a finite number of real polynomial functions $P_{ij}$ ad $Q_{ij}$ for which $S=\cap_j\cup_i\{x\in\mathbb{R}^n:P_{ij}(x)=0,~Q_{ij}(x)<0\}$. A function $f$ is said to be a semialgebraic function if its graph ${(x,y)\in \mathbb{R}^n\times \mathbb{R}:f(x)=y}$ is a semialgebraic set. The subset $S\in \mathbb{R}$ is a semialgebraic set if there exists a finite number of real polynomial functions $P_{ij}$ ad $Q_{ij}$ for which $S=\cap_j\cup_i\{x\in\mathbb{R}^n:P_{ij}(x)=0,~Q_{ij}(x)<0\}$. A function $f$ is said to be a semialgebraic function if its graph ${(x,y)\in \mathbb{R}^n\times \mathbb{R}:f(x)=y}$ is a semialgebraic set.
\end{definition}
\begin{remark}
It is a fact that the semialgebraic real-valued function has (K-Ł) property at any point $\bar x\in dom(\partial f)$; i.e., they are K-Ł functions.
\end{remark}
With the above prerequisites in place, we can now proceed with the proof. For convenience, let $\mathcal{D}=\{\mathcal{X,Z,C},\mathbf{G,H,T}\}$ and $y(\mathcal{D})=\sum_{i=1}^{n_3}\Vert{\mathcal{Z}}^{(i)}\Vert_*+\frac{\alpha}{2} \Vert\mathcal{C}-\mathcal{X}\bar{ \star}_{(1,2)} \mathbf{G}\star_{(1,3)} \mathbf{H}\times_3 \mathbf{T}\Vert_F^2+\frac{\beta}{2}\Vert \mathcal{Z}-\psi(\mathcal{C})\Vert_F^2 + \Phi(\mathcal{X})+\Upsilon{(\mathbf{G})}+\Upsilon{(\mathbf{H})}+\Upsilon{(\mathbf{T})}$, $y^\prime (\mathcal{D})=\frac{\alpha}{2} \Vert\mathcal{C}-\mathcal{X}\bar{ \star}_{(1,2)} \mathbf{G}\star_{(1,3)} \mathbf{H}\times_3 \mathbf{T}\Vert_F^2+\frac{\beta}{2}\Vert \mathcal{Z}-\psi(\mathcal{C})\Vert_F^2$.
\begin{lemma}[Sufficient decrease]
\label{lemma:sufficient}
Assume that the sequence $\{\mathcal{X}^n,\mathcal{Z}^n, \mathcal{C}^n,\mathbf{G}^n,\mathbf{H}^n,\mathbf{T}^n\}_{n=1}^{+\infty}$ is yielded by the updating process Eq. (\ref{eq:process}) of designed algorithm. And denote $\mathcal{D}^k=\{\mathcal{X}^k,\mathcal{Z}^k, \mathcal{C}^k,\mathbf{G}^k,\mathbf{H}^k,\mathbf{T}^k\}$, then we have the following conclusion,
\begin{equation*}
y(\mathcal{D}^{k+1})+\rho\|\mathcal{D}^{k+1}-\mathcal{D}^k\|_F^2\le y(\mathcal{D}^k)
\end{equation*}
where $\rho=\frac{1}{2}\min_{1\le i\le 6}{\rho_i}$ .  
\end{lemma}

\textit{proof.} Let $\mathcal{X}^{k+1},\mathcal{Z}^{k+1}, \mathcal{C}^{k+1},\mathbf{G}^{k+1},\mathbf{H}^{k+1},\mathbf{T}^{k+1}$ be the optimal solutions with respect to each subproblem described in Section \ref{subproblems}, then the following inequalities hold:		
% \vskip -0.35in
\begin{equation*}
\begin{cases}
y(\mathcal{X}^{k+1},\mathcal{Z}^{k+1}, \mathcal{C}^k,\mathbf{G}^k,\mathbf{H}^k,\mathbf{T}^k)+\frac{\rho_2}{2}\|\mathcal{Z}^{k+1}-\mathcal{Z}^k\|^2\le \\y(\mathcal{X}^{k+1},\mathcal{Z}^k, \mathcal{C}^k,\mathbf{G}^k,\mathbf{H}^k,\mathbf{T}^k),
\\y(\mathcal{X}^{k+1},\mathcal{Z}^{k+1}, \mathcal{C}^{k+1},\mathbf{G}^k,\mathbf{H}^k,\mathbf{T}^k)+\frac{\rho_3}{2}\|\mathcal{C}^{k+1}-\mathcal{C}^k\|^2\le \\y(\mathcal{X}^{k+1},\mathcal{Z}^{k+1}, \mathcal{C}^k,\mathbf{G}^k,\mathbf{H}^k,\mathbf{T}^k),\\
~~~~~~~~~~~~~~~~~\vdots
% y(\mathcal{X}^{k+1},\mathcal{Z}^{k+1}, \mathcal{C}^{k+1},\mathbf{G}^{k+1},\mathbf{H}^k,\mathbf{T}^k)\\
% +\frac{\rho_4}{2}\|\mathbf{G}^{k+1}-\mathbf{G}^k\|^2\le y(\mathcal{X}^{k+1},\mathcal{Z}^{k+1},\mathcal{C}^{k+1},\mathbf{G}^k,\mathbf{H}^k,\mathbf{T}^k)
% \\y(\mathcal{X}^{k+1},\mathcal{Z}^{k+1}, \mathcal{C}^{k+1},\mathbf{G}^{k+1},\mathbf{H}^{k+1},\mathbf{T}^k)\\
% +\frac{\rho_5}{2}\|\mathbf{H}^{k+1}-\mathbf{H}^k\|^2\le y(\mathcal{X}^{k+1},\cdots,\mathbf{H}^k,\mathbf{T}^k)
\\y(\mathcal{X}^{k+1},\mathcal{Z}^{k+1}, \mathcal{C}^{k+1},\mathbf{G}^{k+1},\mathbf{H}^{k+1},\mathbf{T}^{k+1})\\
+\frac{\rho_6}{2}\|\mathbf{T}^{k+1}-\mathbf{T}^k\|^2\le y(\mathcal{X}^{k+1},\mathcal{Z}^{k+1}, \cdots,\mathbf{H}^{k+1},\mathbf{T}^k)
\end{cases}
\end{equation*} Combining the above inequalities, we get
\begin{align*}
&y(\mathcal X^{k+1},Z^{k+1}, C^{k+1},\mathbf{G}^{k+1},\mathbf{H}^{k+1},\mathbf{T}^{k+1})\\&+\frac{\rho_6}{2}\|\mathbf{T}^{k+1}-\mathbf{T}^k\|^2+\frac{\rho_5}{2}\|\mathbf{H}^{k+1}-\mathbf{H}^k\|^2 +\\&\frac{\rho_4}{2}\|\mathbf{G}^{k+1}-\mathbf{G}^k\|^2+\frac{\rho_3}{2}\|\mathcal{C}^{k+1}-\mathcal{C}^k\|^2+\frac{\rho_2}{2}\|\mathcal{Z}^{k+1}-\mathcal{Z}^k\|^2\\&+\frac{\rho_1}{2}\|\mathcal{X}^{k+1}-\mathcal{X}^k\|^2\\
&\le y(\mathcal{X}^{k},\mathcal{Z}^{k}, \mathcal{C}^{k},\mathbf{G}^{k},\mathbf{H}^{k},\mathbf{T}^{k})
\end{align*}
i.e.,
\begin{equation*}
y(\mathcal{D}^{k+1})+\rho\|\mathcal{D}^{k+1}-\mathcal{D}^k\|_F^2 \le y(\mathcal{D}^k)
\end{equation*}
with $\rho=\frac{1}{2}\min_{1\le i\le 6}{\rho_i}$, as desired.

\begin{lemma}[relative error lemma]
\label{lemma:relative}
Let $\mathcal{D}^k=\{\mathcal{}X^k,\mathcal{Z}^k, \mathcal{C}^k,\mathbf{G}^k,\mathbf{H}^k,\mathbf{T}^k\}$ be the sequence generated by the process of the designed algorithm, then $\mathcal{D}^k$ is bounded for all natural number $k$, and there exists $e^{k+1}\in\partial y(D^{k+1})$ such that
\begin{equation*}
\|e^{k+1}\|_F^2 \le c\|D^{k+1}-D^k\|_F^2
\end{equation*}
where $c=\max_{1\le i\le6}|\rho_i|+\theta$ in which $\theta$ is the Lipschitz constant of $~\nabla y^\prime$. 
\end{lemma}

\textit{proof.} Assume for the sake of contradiction that for any natural number $k$, $\mathcal D^k$ is unbounded, which means that there exist some elements in $\mathcal D^k$ whose Frobenius norms approach $\infty$.
Since
\begin{align*}
&\lim_{\|\mathcal{X}\|_F\to \infty}\frac{\alpha}{2} \Vert\mathcal{C}-\mathcal{X}\bar{ \star}_{(1,2)} \mathbf{G}\star_{(1,3)} \mathbf{H}\times_3 \mathbf{T}\Vert_F^2=+\infty,
\\&\lim_{\|\mathcal{Z}\|_F\to\infty}\frac{\beta}{2}\Vert \mathcal{Z}-\psi(\mathcal{C})\Vert_F^2=+\infty,
\\&\lim_{\|\mathcal{C}\|_F\to \infty}\frac{\alpha}{2} \Vert\mathcal{C}-\mathcal{X}\bar{ \star}_{(1,2)} \mathbf{G}\star_{(1,3)} \mathbf{H}\times_3 \mathbf{T}\Vert_F^2=+\infty,
\\&\lim_{\|\mathbf{G}\|_F\to\infty}\frac{\alpha}{2} \Vert\mathcal{C}-\mathcal{X}\bar{ \star}_{(1,2)} \mathbf{G}\star_{(1,3)} \mathbf{H}\times_3 \mathbf{T}\Vert_F^2=+\infty,
\\&\lim_{\|\mathbf{H}\|_F\to\infty}\frac{\alpha}{2} \Vert\mathcal{C}-\mathcal{X}\bar{ \star}_{(1,2)} \mathbf{G}\star_{(1,3)} \mathbf{H}\times_3 \mathbf{T}\Vert_F^2=+\infty,
\\&\lim_{\|\mathbf{T}\|_F\to\infty}\frac{\alpha}{2} \Vert\mathcal{C}-\mathcal{X}\bar{ \star}_{(1,2)} \mathbf{G}\star_{(1,3)} \mathbf{H}\times_3 \mathbf{T}\Vert_F^2=+\infty.
\end{align*} thus clearly, it is true that
\begin{align*}
&\lim_{\|\mathcal{X}\|_F\to\infty}y(\mathcal D)=+\infty,~~\lim_{\|\mathcal{Z}\|_F\to\infty}y(\mathcal D)=+\infty, \\
&\lim_{\|\mathcal{C}\|_F\to\infty}y(\mathcal D)=+\infty,~~\lim_{\|\mathbf G\|_F\to\infty}y(\mathcal D)=+\infty,\\
&\lim_{\|\mathbf{H}\|_F\to\infty}y(\mathcal D)=+\infty,~~\lim_{\|\mathbf{T}\|_F\to\infty}y(\mathcal D)=+\infty.
\end{align*} But by lemma \ref{lemma:sufficient}, for every positive natural number $k$, we have 
\begin{equation*}
y(\mathcal D^{k})\le y(D^{k})+\rho\|\mathcal D^{k}-D^{k-1}\|_F^2\le y(\mathcal D^{k-1})\le\dots\le y(\mathcal{D}^0)
\end{equation*}
in which the last term $y(\mathcal{D}^0)$ is a bounded real value so that $y(\mathcal{D}^k)<+\infty$, a contradiction. Thus, $D^k$ is bounded for all natural numbers $k$.
Let \{$\mathcal{X}^{k+1},\mathcal{Z}^{k+1}, \mathcal{C}^{k+1},\mathbf{G}^{k+1},\mathbf{H}^{k+1},\mathbf{T}^{k+1}\}$ be the optimal solutions of each subproblem in \ref{subproblems}, according to the Karush-Kuhn-Tucker conditions, for any step $k+1$, we have the following formulas
\begin{align*}
&0\in\partial(\Phi(\mathcal{X}^{k+1})+\frac{\alpha}{2} \Vert\mathcal{C}^{k}-\mathcal{X}^{k+1}\bar{ \star}_{(1,2)} \mathbf{G}^k\star_{(1,3)} \mathbf{H}^k\times_3 \mathbf{T}^k\Vert_F^2
\\&+\frac{\rho_1}{2}\Vert\mathcal{X}^{k+1}-\mathcal{X}^{k}\Vert_F^2),\\&0\in\partial(\Vert\mathcal{Z}^{k+1}\Vert_*+\frac{\beta}{2}\Vert\mathcal{Z}^{k+1}-\psi(\mathcal{C}^k)\Vert_F^2 + \frac{\rho_2}{2}\Vert \mathcal{Z}^{k+1}-\mathcal{Z}^k\Vert_F^2),\\
&0=\partial(\frac{\alpha}{2}\Vert\mathcal{C}^{k+1}-\mathcal{X}^k\bar{ \star}_{(1,2)} \mathbf{G}^k\star_{(1,3)} \mathbf{H}^k\times_3 \mathbf{T}^k\Vert_F^2 + \frac{\beta}{2}\Vert \mathcal{Z}^k-
\\&\psi(\mathcal{C}^{k+1})\Vert_F^2 + \frac{\rho_3}{2}\Vert\mathcal{C}^{k+1}-\mathcal{C}^k\Vert_F^2),\\
&0\in\partial(\Upsilon(\mathbf{G}^{k+1}) + \frac{\alpha}{2}\Vert\mathcal{C}^k-\mathcal{X}^k\bar{ \star}_{(1,2)} \mathbf{G}^{k+1}\star_{(1,3)} \mathbf{H}^k\times_3 \mathbf{T}^k\Vert_F^2
\\& + \frac{\rho_4}{2}\Vert \mathbf{G}^{k+1}-\mathbf{G}^k\Vert_F^2),
\\&0\in\partial(\Upsilon(\mathbf{H}^{k+1}) + \frac{\alpha}{2}\Vert\mathcal{C}^k-\mathcal{X}^k\bar{ \star}_{(1,2)} \mathbf{G}^k\star_{(1,3)} \mathbf{H}^{k+1}\times_3 \mathbf{T}^k\Vert_F^2 \\&+ \frac{\rho_5}{2}\Vert \mathbf{H}^{k+1}-\mathbf{H}^k\Vert_F^2),
\\&0\in\partial(\Upsilon(\mathbf{T}^{k+1}) + \frac{\alpha}{2}\Vert\mathcal{C}^k-\mathcal{X}^k\bar{ \star}_{(1,2)} \mathbf{G}^k\star_{(1,3)} \mathbf{H}^k\times_3 \mathbf{T}^{k+1}\Vert_F^2
\\& + \frac{\rho_6}{2}\Vert \mathbf{T}^{k+1}-\mathbf{T}^k\Vert_F^2).
\end{align*}
By introducing
\begin{align*}
&V_1^{k+1}=-\partial(\frac{\alpha}{2} \Vert\mathcal{C}^{k}-\mathcal{X}^{k+1}\bar{ \star}_{(1,2)} \mathbf{G}^k\star_{(1,3)} \mathbf{H}^k\times_3 \mathbf{T}^k\Vert_F^2\\&+\frac{\rho_1}{2}\Vert\mathcal{X}^{k+1}-\mathcal{X}^{k}\Vert_F^2),\\
&V_2^{k+1}=-\partial(\frac{\beta}{2}\Vert\mathcal{Z}^{k+1}-\psi(\mathcal{C}^k) \Vert_F^2 + \frac{\rho_2}{2}\Vert \mathcal{Z}^{k+1}-\mathcal{Z}^k\Vert_F^2),\\
&V_3^{k+1}=0,\\
&V_4^{k+1}=-\partial(\frac{\alpha}{2}\Vert\mathcal{C}^k-\mathcal{X}^k\bar{ \star}_{(1,2)} \mathbf{G}^{k+1}\star_{(1,3)} \mathbf{H}^k\times_3 \mathbf{T}^k\Vert_F^2 \\&+ \frac{\rho_4}{2}\Vert \mathbf{G}^{k+1}-\mathbf{G}^k\Vert_F^2),\\
&V_5^{k+1}=-\partial(\frac{\alpha}{2}\Vert\mathcal{C}^k-\mathcal{X}^k\bar{\star}_{(1,2)} \mathbf{G}^k\star_{(1,3)} \mathbf{H}^{k+1}\times_3 \mathbf{T}^k\Vert_F^2 \\&+ \frac{\rho_5}{2}\Vert \mathbf{H}^{k+1}-\mathbf{H}^k\Vert_F^2),\\
&V_6^{k+1}=-\partial(\frac{\alpha}{2}\Vert\mathcal{C}^k-\mathcal{X}^k\bar{\star}_{(1,2)} \mathbf{G}^k\star_{(1,3)} \mathbf{H}^k\times_3 \mathbf{T}^{k+1}\Vert_F^2 \\&+ \frac{\rho_6}{2}\Vert \mathbf{T}^{k+1}-\mathbf{T}^k\Vert_F^2).
\end{align*}it is obvious that we have 
\begin{align*}
&\|V_1^{k+1}+\nabla_{\mathcal X}y^\prime(\mathcal{X}^{k+1},\mathcal{Z}^{k}, \mathcal{C}^{k},\mathbf{G}^{k},\mathbf{H}^{k},\mathbf{T}^{k})\|_F^2\\&\le \rho_1\|\mathcal{X}^{k+1}-\mathcal{X}^k\|_F^2, \\
&\|V_2^{k+1}+\nabla_{\mathcal Z} y^\prime(\mathcal{X}^{k},\mathcal{Z}^{k+1}, \mathcal{C}^{k},\mathbf{G}^{k},\mathbf{H}^{k},\mathbf{T}^{k})\|_F^2\\&\le \rho_2\|\mathcal{Z}^{k+1}-\mathcal{Z}^k\|_F^2, \\
% &\|V_3^{k+1}+\nabla_{\mathcal C} y^\prime(\mathcal{X}^{k},\mathcal{Z}^{k}, \mathcal{C}^{k+1},\mathbf{G}^{k},\mathbf{H}^{k},\mathbf{T}^{k})\|_F^2\\&\le \rho_3\|\mathcal{C}^{k+1}-\mathcal{C}^k\|_F^2, \\
% &\|V_4^{k+1}+\nabla_{\mathbf G} y^\prime(\mathcal{X}^{k},\mathcal{Z}^{k}, \mathcal{C}^{k},\mathbf{G}^{k+1},\mathbf{H}^{k},\mathbf{T}^{k})\|_F^2\\&\le \rho_4\|\mathbf{G}^{k+1}-\mathbf{G}^k\|_F^2,\\ 
% &\|V_5^{k+1}+\nabla_{\mathbf H} y^\prime(\mathcal{X}^{k},\mathcal{Z}^{k}, \mathcal{C}^{k},\mathbf{G}^{k},\mathbf{H}^{k+1},\mathbf{T}^{k})\|_F^2\\&\le \rho_5\|\mathbf{H}^{k+1}-\mathbf{H}^k\|_F^2, \\
&~~~~~~~~~~~~~~~~~\vdots\\
&\|V_6^{k+1}+\nabla_{\mathbf T} y^\prime(\mathcal{X}^{k},\mathcal{Z}^{k}, \mathcal{C}^{k},\mathbf{G}^{k},\mathbf{H}^{k},\mathbf{T}^{k+1})\|_F^2\\&\le \rho_6\|\mathbf{T}^{k+1}-\mathbf{T}^k\|_F^2. 
\end{align*}
then we define $e^{k+1}=\{e_1^{k+1},e_2^{k+1},e_3^{k+1},e_4^{k+1},e_5^{k+1},e_6^{k+1}\}$ where
\begin{align*}
&e_1^{k+1}=V_1^{k+1}+\nabla_{\mathcal X}y^\prime(\mathcal{X}^{k+1},\mathcal{Z}^{k}, \mathcal{C}^{k},\mathbf{G}^{k},\mathbf{H}^{k},\mathbf{T}^{k}),\\
&e_2^{k+1}=V_2^{k+1}+\nabla_{\mathcal Z}y^\prime(\mathcal{X}^{k},\mathcal{Z}^{k+1}, \mathcal{C}^{k},\mathbf{G}^{k},\mathbf{H}^{k},\mathbf{T}^{k}),\\
% &e_3^{k+1}=V_3^{k+1}+\nabla_{\mathcal C}y^\prime(\mathcal{X}^{k},\mathcal{Z}^{k}, \mathcal{C}^{k+1},\mathbf{G}^{k},\mathbf{H}^{k},\mathbf{T}^{k}),\\
% &e_4^{k+1}=V_4^{k+1}+\nabla_{\mathbf G}y^\prime(\mathcal{X}^{k},\mathcal{Z}^{k}, \mathcal{C}^{k},\mathbf{G}^{k+1},\mathbf{H}^{k},\mathbf{T}^{k}),\\
% &e_5^{k+1}=V_5^{k+1}+\nabla_{\mathbf H}y^\prime(\mathcal{X}^{k},\mathcal{Z}^{k}, \mathcal{C}^{k},\mathbf{G}^{k},\mathbf{H}^{k+1},\mathbf{T}^{k}),\\
&~~~~~~~~~~~~~~~~~\vdots\\
&e_6^{k+1}=V_6^{k+1}+\nabla_{\mathbf T}y^\prime(\mathcal{X}^{k},\mathcal{Z}^{k}, \mathcal{C}^{k},\mathbf{G}^{k},\mathbf{H}^{k},\mathbf{T}^{k+1}).
\end{align*}
Combining the boundedness of $\mathcal{D}^k$ and Lipschitz continuity of $\nabla y^\prime$, we can obtain
\begin{equation*}
\|e^{k+1}\|_F^2 \le c\|D^{k+1}-D^k\|_F^2
\end{equation*}
where $e^{k+1}\in \partial y(\mathcal{D}^{k+1})$, $c=\max_{1\le i\le 6}|\rho_i|+\theta$ and $\theta$ is a Lipschitz constant. The proof is complete.
\begin{theorem}[sequence convergence]
The iterative sequence $\{\mathcal{X}^k,\mathcal{Z}^k, 
\mathcal{C}^k,\mathbf{G}^k,\mathbf{H}^k,\mathbf{T}^k\}_{k=1}^{+\infty}$ obtained by the designed PAM algorithm converges to a critical point of $y({\mathcal{X,Z, C},\mathbf{G},\mathbf{H},\mathbf{T}})$.
\end{theorem}

\textit{proof.} By Theorem 6.2 in \cite{attouch2013convergence}, in order to prove that the sequence $\{\mathcal{X}^k,\mathcal{Z}^k, \mathcal{C}^k,\mathbf{G}^k,\mathbf{H}^k,\mathbf{T}^k\}_{k=1}^{+\infty}$ converges to a critical point of $y$, it suffices to verify the following three conditions:

1. $y(\mathcal{X,Z,C},\mathbf{G}^k,\mathbf{H}^k,\mathbf{T}^k)$ is a proper lower semicontinuous function;

2. $y(\mathcal{X,Z,C},\mathbf{G}^k,\mathbf{H}^k,\mathbf{T}^k)$ is a K-Ł function;

3. The sequence $\{\mathcal{X}^k,\mathcal{Z}^k, \mathcal{C}^k,\mathbf{G}^k,\mathbf{H}^k,\mathbf{T}^k\}_{k=1}^{+\infty}$ satisfies the sufficient decrease and relative error properties.

By lemma \ref{lemma:sufficient} and lemma \ref{lemma:relative}, the third condition has already been met; so it remains to verify the first and second conditions. We illustrate them one by one.

First, $y^\prime({\mathcal{X,Z,C},\mathbf{G},\mathbf{H},\mathbf{T}})$ is a $C^1$ function with local Lipschitz continuous gradients, and the indicator functions $\Phi(\mathcal{X})$, $\Upsilon(\mathbf{G})$, $\Upsilon(\mathbf{H})$, $\Upsilon(\mathbf{T})$ and the nuclear norm term $\sum_{i=1}^{n_3}\|\mathcal{Z}^{(i)}\Vert_*$ are proper lower semicontinuous. Thus $y({\mathcal{X,Z,C},\mathbf{G},\mathbf{H},\mathbf{T}})$ is a proper lower semicontinuous function.

Second, since a semialgebraic real-valued function is a K-Ł function, we turn to check each part of $y$ is a semialgebraic real-valued function.

(1). The term of matrix nuclear norm $\sum_{i=1}^{n_3}\|\mathcal{Z}^{(i)}\Vert_*$ is a semialgebraic real-valued function, see \cite{bolte2014proximal};

(2). The function $y^\prime$ containing only Frobenius norms is a semialgebraic real-valued function, see \cite{bolte2014proximal};

(3). $\Phi(\mathcal{X})$, $\Upsilon(\mathbf{G})$, $\Upsilon(\mathbf{H})$, $\Upsilon(\mathbf{T})$ are indicator functions with semialgebaric sets, see \cite{bolte2014proximal}.

Thus $y$ is a semialgebraic real-valued function; indeed, it is a  K-Ł function.

Hence, the theorem is proved, and we conclude that the sequence $\{\mathcal{X}^k,\mathcal{Z}^k, \mathcal{C}^k,\mathbf{G}^k,\mathbf{H}^k,\mathbf{T}^k\}_{k=1}^{+\infty}$  yielded by the designed PAM algorithm converges to a critical point of the function $y({\mathcal{X Z, C},\mathbf{G},\mathbf{H},\mathbf{T}})$.

\end{document}